\documentclass[preprint,12pt,authoryear,nopreprintline]{elsarticle}

\usepackage{amssymb}
\usepackage{amsmath}
\usepackage{units}
\usepackage{tabularx}

\usepackage{color}

\usepackage{lineno}

\usepackage[colorlinks]{hyperref}
\usepackage{cleveref}

\Crefname{figure}{Fig.}{Figs.}
\Crefname{equation}{Eq.}{Eqs.}
\Crefname{table}{Tab.}{Tabs.}
\Crefname{section}{Sec.}{Secs.}

\journal{JOV: Uncertainty in Sensory Processing and Action Control}

\begin{document}
\begin{frontmatter}

\title{A Robotics-Inspired Scanpath Model Reveals the Importance of Uncertainty and Semantic Object Cues for Gaze Guidance in Dynamic Scenes}


\author[inst1,inst3]{Vito Mengers\corref{sameContrib}}
\author[inst1,inst3]{Nicolas Roth\corref{sameContrib}}
\author[inst1,inst3]{Oliver Brock\corref{sameSuperv}}
\author[inst1,inst3]{Klaus Obermayer\corref{sameSuperv}}
\author[inst2,inst3]{Martin Rolfs\corref{sameSuperv}}

\cortext[sameContrib]{These authors contributed equally. Correspondence to: v.mengers@tu-berlin.de roth@tu-berlin.de}
\cortext[sameSuperv]{Equal supervision.}
\affiliation[inst1]{organization={Technische Universität Berlin},
            addressline={Marchstraße 23}, 
            city={Berlin},
            postcode={10587}, 
            country={Germany}}
\affiliation[inst2]{organization={Humboldt-Universtät zu Berlin},
            addressline={Rudower Chaussee 18}, 
            city={Berlin},
            postcode={12489}, 
            country={Germany}}
\affiliation[inst3]{organization={Science of Intelligence, Research Cluster of Excellence},
            addressline={Marchstraße 23}, 
            city={Berlin},
            postcode={10587}, 
            country={Germany}}

\begin{abstract}
The objects we perceive guide our eye movements when observing real-world dynamic scenes. Yet, gaze shifts and selective attention are critical for perceiving details and refining object boundaries. Object segmentation and gaze behavior are, however, typically treated as two independent processes. Here, we present a computational model that simulates these processes in an interconnected manner and allows for hypothesis-driven investigations of distinct attentional mechanisms. Drawing on an information processing pattern from robotics, we use a Bayesian filter to recursively segment the scene, which also provides an uncertainty estimate for the object boundaries that we use to guide active scene exploration. We demonstrate that this model closely resembles observers' free viewing behavior on a dataset of dynamic real-world scenes, measured by scanpath statistics, including foveation duration and saccade amplitude distributions used for parameter fitting and higher-level statistics not used for fitting. These include how object detections, inspections, and returns are balanced and a delay of returning saccades without an explicit implementation of such temporal inhibition of return. Extensive simulations and ablation studies show that uncertainty promotes balanced exploration and that semantic object cues are crucial to forming the perceptual units used in object-based attention. Moreover, we show how our model's modular design allows for extensions, such as incorporating saccadic momentum or pre-saccadic attention, to further align its output with human scanpaths.


\end{abstract}

\begin{keyword}
scanpath simulation \sep object-based attention \sep probabilistic image segmentation \sep active interconnect \sep eye movement \sep active vision


\end{keyword}

\end{frontmatter}

\section{Introduction}
\label{sec:intro}

Humans actively move their eyes to pay attention to individual parts of their environment. Several seminal studies have explored eye movements in natural contexts \citep{land1994we, land1999roles, pelz2001coordination, triesch2003you, rothkopf2007task, tatler2011eye, mital2011clustering, matthis2018gaze}, yet we lack a mechanistic understanding of gaze control in such natural conditions. Computational models of visual attention provide an invaluable tool to analyze the contributions of distinct mechanisms and link them to observable behavior \citep{itti2001computational, borji2012state, roth2023objects, kummerer2023predicting}.
In this work, we present an object-based computational model that reproduces human free-viewing eye-tracking data (with a stationary head position) when observing natural dynamic scenes. 
In addition to the saccadic decision-making process, we also model how the basic building blocks---on which object-based attention can act---can be formed. Our model is mechanistic in the sense that it implements algorithmic principles behind attentional mechanisms. Specifically, we aim to capture how information is integrated to determine the next saccade target and how different object cues contribute to the formation of perceptual units for object-based attention. However, we do not prioritize the plausibility of how the inputs to these mechanisms are computed in the first place, nor do we make claims about the neural implementation of these mechanisms in the brain. The model's modularity then allows us to systematically test the effect and contribution of different attentional mechanisms on the simulated gaze behavior, which can be directly compared with human eye-tracking results.

Visual attention sequentially selects objects for perceptual processing and provides the information to generate a motor plan for eye movements
\citep{deubel1996saccade}. Different psychophysical experiments have, depending on the task and presented visual stimulus, uncovered different aspects of attention (for an overview, see \citealp{carrasco2011visual, nobre2014oxford}).
The most prominent theories of visual attention describe it as space-, feature-, or object-based. Space-based attention is classically characterized as a spotlight \citep{posner1980orienting} or zoom lens \citep{eriksen1985allocation} that enhances processing at the attended location. The attended location is typically selected based on maxima in a priority or saliency map \citep{koch1985shifts, itti2001computational}. Independent of a specific location, feature-based attention can be deployed covertly to objects that share a specific attribute, like color or motion direction \citep{saenz2002global, treue1999feature, white2011feature}. Evidence for object-based attention was, for example, found in experiments where attention was allocated to one of two objects that share the same location \citep{duncan1984selective, oCraven1999fmri, blaser2000tracking} and where attention was directed faster to locations within an attended object than to locations outside the object \citep{egly1994shifting, malcolm2015object}. The object-specificity of attention suggests that, at least in some cases, the underlying units of attentional processing and selection are discrete visual objects (for reviews, see \citealp{scholl2001objects, peters2021capturing}). \citet{cavanagh2023architecture} presented a compelling framework for how experimental findings attributed to space- or feature-based attention can be conceptualized as forms of object-based attention. We have previously demonstrated using a computational modeling approach that objects are particularly important for gaze guidance during free viewing of dynamic natural scenes \citep{roth2023objects}.

When simulating human eye movements in natural scenes, models are typically limited in at least one of two ways: modeling only the average spatial gaze density instead of individual scanpaths, or being only applicable to static images instead of videos. Classic saliency models have been extended to include motion (e.g., \citealp{molin2015motion}), and deep learning models have been used successfully for video saliency prediction (e.g., \citealp{wang2018revisiting, droste2020unified}). However, saliency models are restricted to modeling the average spatial distribution of gaze positions. Models capable of describing the attentional dynamics of individual saccadic decisions usually assume the scene to be static and are not applicable to dynamic scenes (e.g., \citealp{itti1998model, tatler2017latest, wloka2018active, schwetlick2020modeling, kummerer2022deepgaze}). Rather than relying on these simplifications of common models (for reviews, see \citealp{borji2012state, bylinskii2015towards,kummerer2023predicting}), our approach predicts full scanpaths, including the order and timing of fixation and smooth pursuit events, for dynamic videos. Our previous scanpath model \citep{roth2023objects} describes the saccadic decision-making processes during the free-viewing of dynamic scenes but requires explicitly provided object segmentations for modeling object-based attention. How the building blocks of object-based attention arise before being actively attended and what mechanisms contribute to the formation of these perceptual units are, however, open questions \citep{wagemans2012century}.

Classic theories of the visual system propose that visual processing involves organizing elements of the scene into coherent units through structured operations \citep{ullman1984visual, bundesen1990theory}. To describe what object-based attention can act on, ``proto-objects'' were introduced as pre-attentive volatile units that can be accessed and further shaped by selective attention \citep{rensink2000dynamic}.
\cite{walther2006modeling} proposed a model that generates such proto-objects for static scenes based on salient regions defined based on color, edges, and luminance. 
In contrast, psychophysical studies showed that saliency-based proto-objects are less predictive of where people look in real-world scenes than semantically defined objects \citep{nuthmann2010object, pajak2013object}. This suggests that pre-attentive objects can also be formed based on semantics and do not rely solely on low-level saliency. 
In the same vein, human reconstruction of local image regions is controlled by semantic object boundaries, which are constructed within 100~ms of scene viewing \citep{neri2017object, liu2009timing}, while rapid serial visual representation tasks show that scene identification can be even faster \citep{potter2014detecting}. 
Although object boundaries are formed globally, the recognition of individual objects and perceiving their visual details still requires selective attention \citep{wolfe1994guided, wolfe2021guided, henderson2003human, underwood2008attention} and the confidence in information about the foveated objects increases \citep{stewart2022humans}.  

Since perceived objects guide eye movements while gaze shifts influence object perception, the modeling of object-based saccadic decisions requires linking the two interdependent processes.
Such interdependences pose a challenge for many modeling approaches that tend to treat model components as almost independent. A similar challenge exists in robotics, where a robot usually needs to decide on actions given the highly interdependent information from its different sensors~\citep{eppner2016lessons}. 
Therefore, we model the interdependent segmentation and saccadic decision-making by employing an information processing pattern from robotics, called Active InterCONnect (AICON; see \citealp{battaje2024aicon}), which has been applied to robustly solve such problems for real-world robotic systems~\citep{martin2022coupled}. It is centered around building bidirectional connections between components that allow for the interpretation of sensory cues while taking into account the extracted information from other components. In a recent example of this approach, we combined motion and appearance segmentation of objects to disambiguate each cue~\citep{mengers2023combining}. By additionally extracting kinematic object motion constraints from their observed motion, predicting their future motion becomes easier~\citep{martin2022coupled}, which in turn simplifies segmenting them~\citep{mengers2023combining}. While these bidirectional interactions of components are similar to the top-down influence of higher abstractions on low-level visual processing in reverse hierarchy theory \citep{ahissar2004reverse} or interpretation-guided segmentation~\citep{tenenbaum1977experiments}, they become more informative by estimating the uncertainty of each component's extracted information. This way, the information of different components can be weighted in their connections, and the robot can act according to the current uncertainty, e.g., by moving more carefully or actively obtaining more information~\citep{bohg2017interactive}. 

We transfer this idea to the modeling of interdependent segmentation and visual exploration in dynamic real-world scenes: The components for visual target selection and segmentation of the scene are in an active interconnection regulated by uncertainty. Segmented objects can act as uncertain perceptual units for target selection, while moving the gaze toward a particular object can resolve the uncertainty over its segmentation. 
The initial segmentation of a presented scene is estimated globally, meaning that objects that have not yet been foveated are also segmented throughout the visual field (cf. \citealt{neisser1967cognitive}). We build on psychophysical evidence, showing that an initial global scene segmentation can be obtained already within the first fixation based on low-level appearance \citep{schyns1994blobs}, motion \citep{reppas1997representation}, and semantic \citep{neri2017object} object cues. These pre-attentive object boundaries are sequentially refined through high-quality segmentation masks of the actively attended (i.e., foveated) objects \citep{henderson2003human}. 
We treat these different sources of object information as inherently uncertain cues, which we combine in a Bayes filter, a recurrent mechanism that optimally combines the different input sources and updates its compressed representation based on new measurements over time \citep{sarkka2013bayesian}. Similar to the related Bayes filter for object segmentation in robotics \citep{mengers2023combining}, the tracked uncertainty over the segmentation is estimated based on the agreement of its measurements over time.
Thus, this uncertainty describes where the existence or location of boundaries between objects is ambiguous (for more details, see \ref{app:uncertainty}).
Combined with other scene features, like visual saliency, this uncertainty about the object segmentation drives the active exploration of the scene and contributes to the saccadic decision-making process. The high-resolution semantic segmentation of the object at the current gaze position, in turn, provides a high-confidence measurement and updates the object representation in the Bayes filter. This reduces the uncertainty at the current gaze position and encourages further exploration of other parts of the scene.

The automatic generation of an uncertainty map as a result of our object segmentation hence provides us with an advantage over existing mechanistic computational models of visual attention. They typically rely on an explicitly implemented mechanism, called ``inhibition of return'' (IOR), to propel exploration (cf.\ \citealp{itti2001computational, zelinsky2008theory, schwetlick2020modeling, roth2023objects}). IOR as an attentional effect was first described by \cite{posner1984components} as the temporary inhibition of the visual processing of recently attended scene parts. While the initial experiment did not involve eye movements, subsequent studies have found a temporal delay of return saccades (\emph{temporal} IOR, cf.\ \citealp{luke2014dissociating}) and that saccades are spatially biased away from previously attended locations (\emph{spatial} IOR, cf.\ \citealp{klein1999inhibition}). These effects were interpreted as a foraging factor to encourage attentional orientation to previously unexplored parts of the scene \citep{klein1999inhibition, klein2000inhibition}. \cite{itti1998model} hence used IOR as a convenient mechanism to inhibit locations in the saliency map to prevent their model from repeatedly selecting the same most salient location. 
Including this inhibition subsequently became the de-facto standard for mechanistic scanpath models.
However, mounting evidence suggests that IOR effects observed in cueing tasks \citep{posner1984components, tipper1991object} do not play a significant role in gaze behavior under most conditions: Fixation distributions in scene viewing and visual search actually find an increased probability of returns and an absence of spatial IOR \citep{smith2009facilitation, smith2011looking, hooge2005inhibition}. 
The effect of temporal IOR in scene viewing has been explained by \citet{wilming2013saccadic} through ``saccadic momentum'', a general dependency of fixation durations and subsequent relative saccade angles tendency for saccades to continue the trajectory of the last saccade \citep{anderson2008directional, smith2009facilitation}.

In the present work, we propose a mechanistic computational scanpath model that does not rely on active IOR as a mechanism to drive scene exploration. Instead, we used a close interaction between the object segmentation and the saccadic decision-making processes to leverage uncertainty over the object boundaries in the scene to encourage exploration. We show that these interconnected processes lead to human-like gaze behavior for dynamic real-world scenes.
The modular implementation of our model allows for principled hypothesis testing by analyzing the influence of different implementations on the simulated gaze behavior. 
We systematically explore the influence of the object uncertainty on the model scanpaths and find that it leads to an exploration behavior that closely resembles the human data. It even reproduces the temporal IOR effect without the need for an explicit IOR implementation. 
Moreover, we show that access to high-level object information leads to more realistic scanpaths, suggesting that perceptual units of human attention are shaped by semantic knowledge.
Finally, we demonstrate how the model can easily be extended to include additional mechanisms like saccadic momentum and pre-saccadic attention.

\section{Materials and methods}
\label{sec:mandm}
\subsection{A model for interdependent saccadic decisions and object segmentation}
\begin{figure}[t]
    \centering
    \includegraphics[width=1\linewidth]{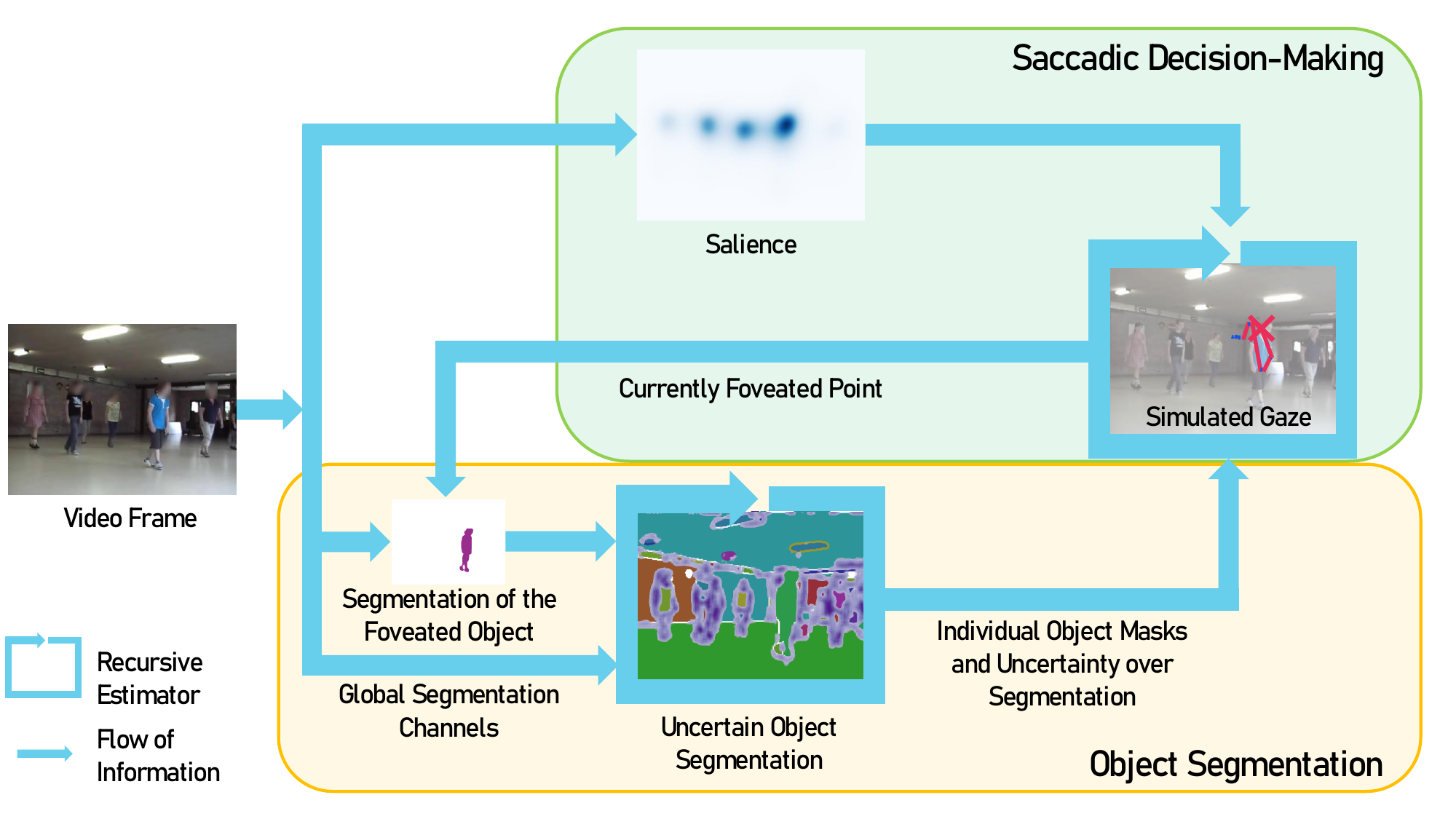}
    \caption{Saccadic decisions and object perception influence each other, as reflected by their interconnection in our model. We illustrate the information flow in our model during the processing of a single frame from a dynamic video. Object segmentation is informed by multiple global object cues and a high-confidence prompted segmentation of the foveated object. The segmented objects act as perceptual units for the saccade target selection. The uncertainty over object segmentation plays a key role in driving exploration while being resolved through high-confidence measurements at the current gaze position. As both the dynamic scene and gaze change over time, the recursive estimator continuously updates the segmentation and its uncertainty.}
    \label{fig:modeloverview}
\end{figure}

We propose a model for the two processes of saccadic decision-making and object segmentation in natural scenes. To establish an active interconnection between them, we employ a design principle from robotics \citep{martin2022coupled} that focuses on bidirectional interactions between components. For our model, this means that we implement both saccade target selection and object segmentation as components that require the other's current state as input, as shown in \Cref{fig:modeloverview}. Critically, we consider the uncertainty of the current segmentation to weigh different segmentation measurements. 
This segmentation uncertainty is also an input to our saccade target selection, as studies of eye movements in natural environments have shown that uncertainty about the state of the visual environment is important to understand and predict gaze behavior \citep{hayhoe2018control, gottlieb2013information}.

Below, we explain how each component models the respective process based on the visual input and the other component's current state. We start with the component for object segmentation, which we adapted from our previous work in robotic perception~\citep{mengers2023combining} to account for object information at the current gaze position and top-down semantic information. Then we explain how we modified our previous model for the saccadic decision-making process~\citep{roth2023objects} to take advantage of both the segmentation and its estimated uncertainty.

\subsubsection{Estimating object segmentation and its uncertainty}
\label{sec:mandm_segmentation}

\begin{figure}[t]
    \centering
    \includegraphics[width=1\linewidth]{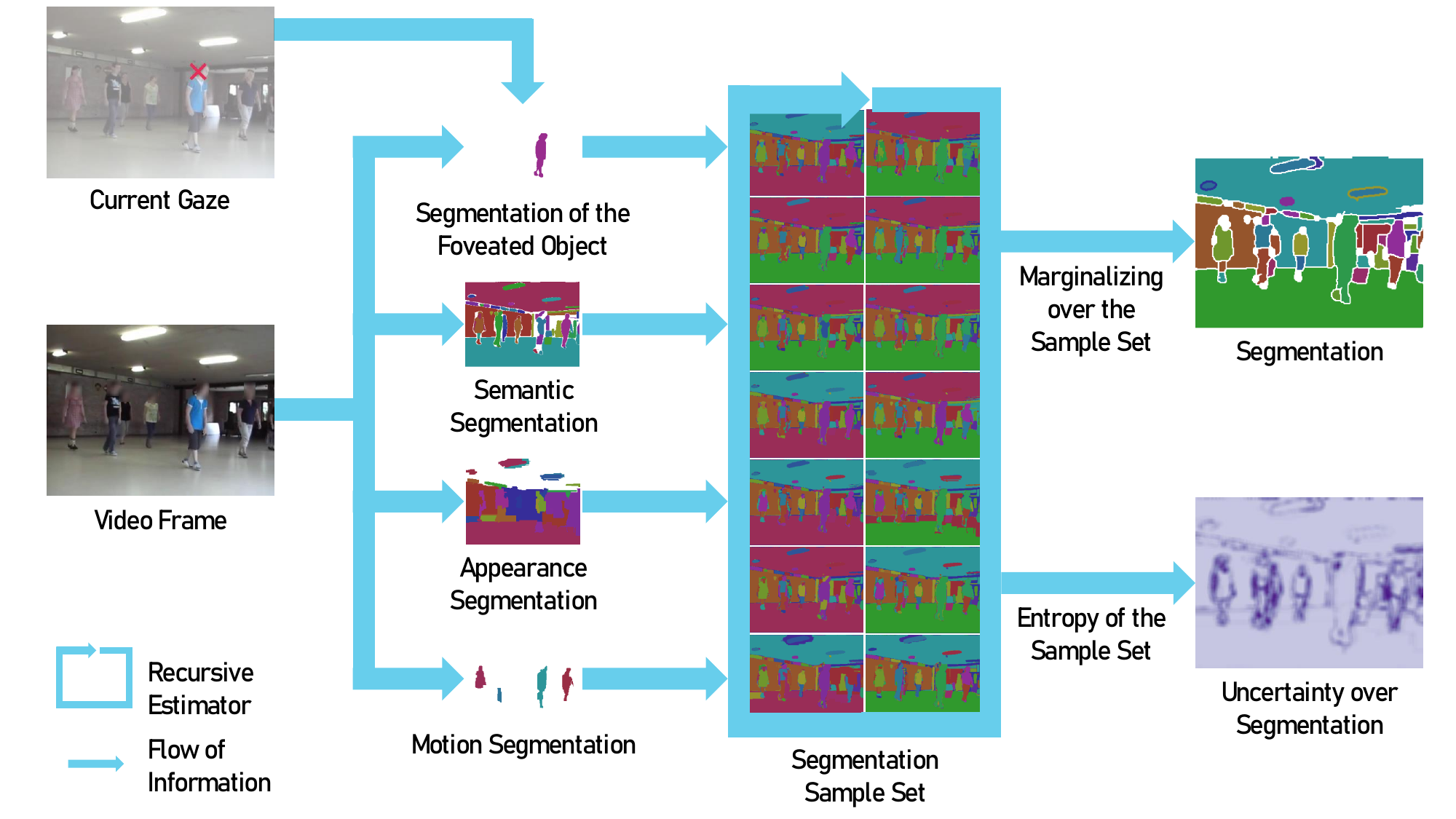}
    \caption{
    Our model combines multiple object cues to estimate both object segmentation and its uncertainty recursively. We integrate foveated and global segmentations of the scene (left) in a Bayesian filter (middle), which maintains a belief over the current state, represented by a weighted set of multiple possible segmentation samples ($14$ example samples from the full set of $50$ are shown). We then compute the currently most likely segmentation and its uncertainty (right), which we use to inform saccadic decisions.}
    \label{fig:object_seg_model}
\end{figure}

In real-world scenes, object segmentations based on semantics, motion, and appearance will typically not wholly agree~\citep{hackett1990multi,pantofaru2008object}. This leads to ambiguity and uncertainty when combining different object cues~\citep{chen1980image,hackett1990multi,pantofaru2008object,mengers2023combining}. For example, an object of similar color to the environment that does not move might be counted towards the background (in \Cref{fig:object_seg_model}, the shirt of the person on the right disappears in appearance segmentation) while an object made up of multiple similarly colored parts that can move relative to another might be segmented into these parts according to their motion (in \Cref{fig:object_seg_model}, the lower half of the person on the left is not moving together with the upper and hence disappears in motion segmentation). 
Therefore, we aim not only to estimate the object segmentation but also to explicitly estimate the current uncertainty over it. To do so, we combine multiple cues for object segmentation as measurements in a recursive Bayesian filter \citep{sarkka2013bayesian}. This filter updates the object segmentation with each new measurement while also estimating its uncertainty, similar to the segmentation filter in previous work on object segmentation for robotics \citep{mengers2023combining}. As shown in \Cref{fig:object_seg_model} on the left, we consider three measurements of pre-attentive global segmentation based on motion, appearance, and semantics, as well as segmentation of only the locally attended object. This attentive segmentation is particularly important because it has higher confidence \citep{stewart2022humans}, thereby reducing segmentation uncertainty dependent on the current gaze. This is one direction of the strong interaction between object segmentation and saccadic decision-making in our model. We now describe how we obtain the different measurements of object segmentation, before explaining how we combine them in a Bayesian way using a particle filter to estimate both segmentation and its uncertainty.

\paragraph*{Cues for the current object segmentation}\label{sec:mandm_segmentation_cues}
We aim to design a directly image-computable model and thus rely only on the RGB video as input for pre-attentive global segmentation. We extract three object segmentation cues from it: low-level appearance, higher-level semantic features, and common motion. For the appearance segmentation, we use the simple graph-based method by \cite{felzenszwalb2004efficient} because it already provides reliable regions of common appearance. For the semantic segmentation, we face a more complex problem, for which we leverage recent advances in large, data-driven segmentation models \citep{kirillov2023sam}. Concretely, we obtain a semantic segmentation using the method by \cite{ke2023samhq}. To find common motion in the scene, we first quantify motion as optical flow between subsequent frames using a state-of-the-art, data-driven method \citep{shi2023videoflow}. We then find parts that move together by applying the same graph-based method \citep{felzenszwalb2004efficient} as for appearance, since it proves to be sufficiently reliable.

Moreover, we use the current gaze location to inform object segmentation because gazing at an object should afford higher-confidence measurements of its boundaries \citep{henderson2003human}. To model such precise measurements around the currently attended object, we use a large data-driven segmentation model \citep{zhao2023fastsam} that can develop a prompted segmentation around a provided point (for more details, see \ref{app:semanticsegmentation}). If we provide it with the current gaze location, we obtain its highest confidence object that contains this point. To further increase the quality of this prompted high-confidence segmentation, we perform it on the highest available resolution of the input image, which we downsample for other cues (see \Cref{tab:params}). We use the prompted segmentation as an additional input to our filter for object segmentation. Since the current gaze location is a result of the previous saccadic decision process, this represents the connection of the two components in one direction. We explain the other, richer direction in \Cref{sec:mandm_scan}~\nameref{sec:mandm_scan}, but now continue to explain how we combine all the described inputs to obtain one object segmentation with uncertainty. 

\paragraph{Combining different object segmentation measurements in a particle filter}

Our aim is to represent object segmentation and its uncertainty, which means a \emph{belief} over object segmentation, and update this belief with new measurements over time. Representing such a belief is hard, because the space of possible segmentations is complex, high-dimensional, and can have multiple modes. Consequently, we cannot simply represent this belief with a Gaussian over object segmentation. We have shown previously that, instead, we can use a Monte-Carlo approach for such representations, where each set of particles corresponds to the likely segmentation of the scene \cite[Sec.~III-A]{mengers2023combining}. These particles together represent a belief over the segmentation, which we can recursively update with a \emph{particle filter} \citep{thrun2005probabilistic}. To give an intuition for this particle filtering approach, let us consider the general problem of estimating a belief over a state $s_\mathrm{t}$ that dynamically changes over time and for which we obtain measurements $z_\mathrm{t}$. When using a particle filter, we represent the belief over the state $s_\mathrm{t}$ by a set of different particles, each a hypothesis $s^{[\mathrm{i}]}$ for the current state. If the state was not dynamic, we could now use the measurements over time to determine the true state by weighting each hypothesis with a weight $w^{[\mathrm{i}]}_\mathrm{t}$ (\Cref{eq:particle_filter_weight} where $\eta$ is a normalizing factor and $i$ is the index of the particle). Unlikely states are removed using weighted resampling, i.e., redetermining the particle set by randomly drawing with replacement particles from the current set according to their weights. To account for dynamism, we can add a prediction step (\Cref{eq:particle_filter_prediction}), where we adapt each hypothesis $s_\mathrm{t}^{[\mathrm{i}]}$ according to available information $a_\mathrm{t}$ on the current development of the state $s_\mathrm{t}$. For a more detailed introduction and derivation of the particle filter, please see \citep{thrun2005probabilistic}. 

\begin{align}
    \forall_i&: w^{[\mathrm{i}]}_\mathrm{t} = \frac{1}{\eta} \cdot p(z_\mathrm{t} | s^{[\mathrm{i}]}_\mathrm{t})\label{eq:particle_filter_weight}\\
    \forall_i&: s^{[\mathrm{i}]}_\mathrm{t} \sim p(s_\mathrm{t} | s^{[\mathrm{i}]}_{\mathrm{t-\Delta t}}, a_\mathrm{t})\label{eq:particle_filter_prediction}
\end{align}

When using such a particle filter to update a belief over the segmentation of a scene, each particle $s^{[\mathrm{i}]}_\mathrm{t}$ is one possible segmentation of the scene into objects (see \Cref{fig:object_seg_model}). Together, these particles represent different hypotheses for the object segmentation of the scene and---in their (dis-)similarities for different parts of the scene---varying levels of uncertainty. We recursively filter this set to account for the dynamism of the scene and integrate new measurements of the real segmentation by implementing \Cref{eq:particle_filter_weight,eq:particle_filter_prediction}: To perform predictions (\Cref{eq:particle_filter_prediction}) of these particles, we use the current optical flow as $a_\mathrm{t}$ to shift the boundaries between objects in each particle's segmentation according to the estimated motion between frames. Then, we weigh the resulting segmentation particles according to their distance to each of our measurements (\Cref{eq:particle_filter_weight}), resampling the set according to the product of the resulting weights. To determine this distance between a particle's segmentation and a measured segmentation, we compute the average distance of their object boundaries, as described in more detail in \ref{app:particle_weighting}. In addition to weighting and resampling the particles based on current segmentation measurements, we also adjust the belief by directly incorporating measured segments into some of the particle segmentations. This is not strictly necessary since these measurements are, in principle, already incorporated in the particles. Still, modifying some of the particles to more closely resemble the current measurements is computationally favorable because it allows for a higher quality of the belief approximation around the most likely areas and makes the approach more robust for a smaller number of particles, as explained in \cite[Sec.~III-C]{mengers2023combining}. The resulting resampled set then represents the currently most likely segmentation hypotheses according to the measurement history.

\paragraph{Obtaining object segmentation and its uncertainty}

While the set of segmentation samples is useful to maintain a belief over the segmentation into objects, it is challenging to utilize in saccadic decision-making. Therefore, we marginalize across the sample set at each time step to obtain one object segmentation and uncertainty estimate, as illustrated on the right in \Cref{fig:object_seg_model}. We first determine the likelihood $p_\mathrm{b}(x,y)$ that each image pixel $(x,y)$ is part of a boundary between two segments by comparing the weights of all particles with a boundary at a given pixel (the particle set $\mathcal{B}(x,y)$) against the weights of those without (the particle set $\Bar{\mathcal{B}}(x,y)$) as shown in \Cref{eq:boundary_likelihood}. Based on these boundary likelihoods, we can then obtain the currently most likely segmentation by thresholding and closing contours. Compared to the full set of segmentation samples, this is, of course, some loss of information, but we preserve the information on the agreement between particles by explicitly deriving the current uncertainty. To do so, we evaluate the entropy $H(x,y)$ of the previously thresholded boundary likelihood (\Cref{eq:entropy}), resulting in high values where some samples have boundaries, while others do not. 

\begin{gather}
    p_\mathrm{b}(x,y) = \frac{\sum_{i \in \,\mathcal{B}(x,y)} \: w^{[\mathrm{i}]}_\mathrm{t} }{\sum_{i \in\, \mathcal{B}(x,y) \,\cup\, \Bar{\mathcal{B}}(x,y)} \: w^{[\mathrm{i}]}_\mathrm{t}}\label{eq:boundary_likelihood}\\
    \begin{split}
            H(x,y) = &- p_\mathrm{b}(x,y) \cdot \log(p_\mathrm{b}(x,y))\\ &- (1 - p_\mathrm{b}(x,y)) \cdot  \log(1 - p_\mathrm{b}(x,y))
    \end{split}\label{eq:entropy}
\end{gather}

We use the obtained object segmentation and uncertainty to select saccade targets in a drift-diffusion model (DDM) over the objects. To do so, we need to ensure that the same object keeps the same ID within the segmentations over time. Therefore, we employ a variation of the Hungarian algorithm~\citep{hopcroft1973} to match object IDs between object segmentations. Specifically, we determine the similarity of the segments in the current object segmentation to those in the past 10 time steps by determining their intersection over union, discounted for older segmentations to favor keeping the currently used object IDs. This results in a weighted bipartite graph from old segment IDs to new segments, in which we find the matching where each new segment is matched with an ID such that the sum of all weights is maximized \citep[see][]{jonker1988shortest}. If no existing ID can be matched, a new ID is assigned. For further details on this matching procedure, see \ref{app:id_matching}.
The segmentation map then forms the basis for the object-based attention mechanism in the scanpath simulation, which we describe in detail in the next section.

\subsubsection{Scanpath simulation}\label{sec:mandm_scan}

\begin{figure}[t]
    \centering
    \includegraphics[width=1\linewidth]{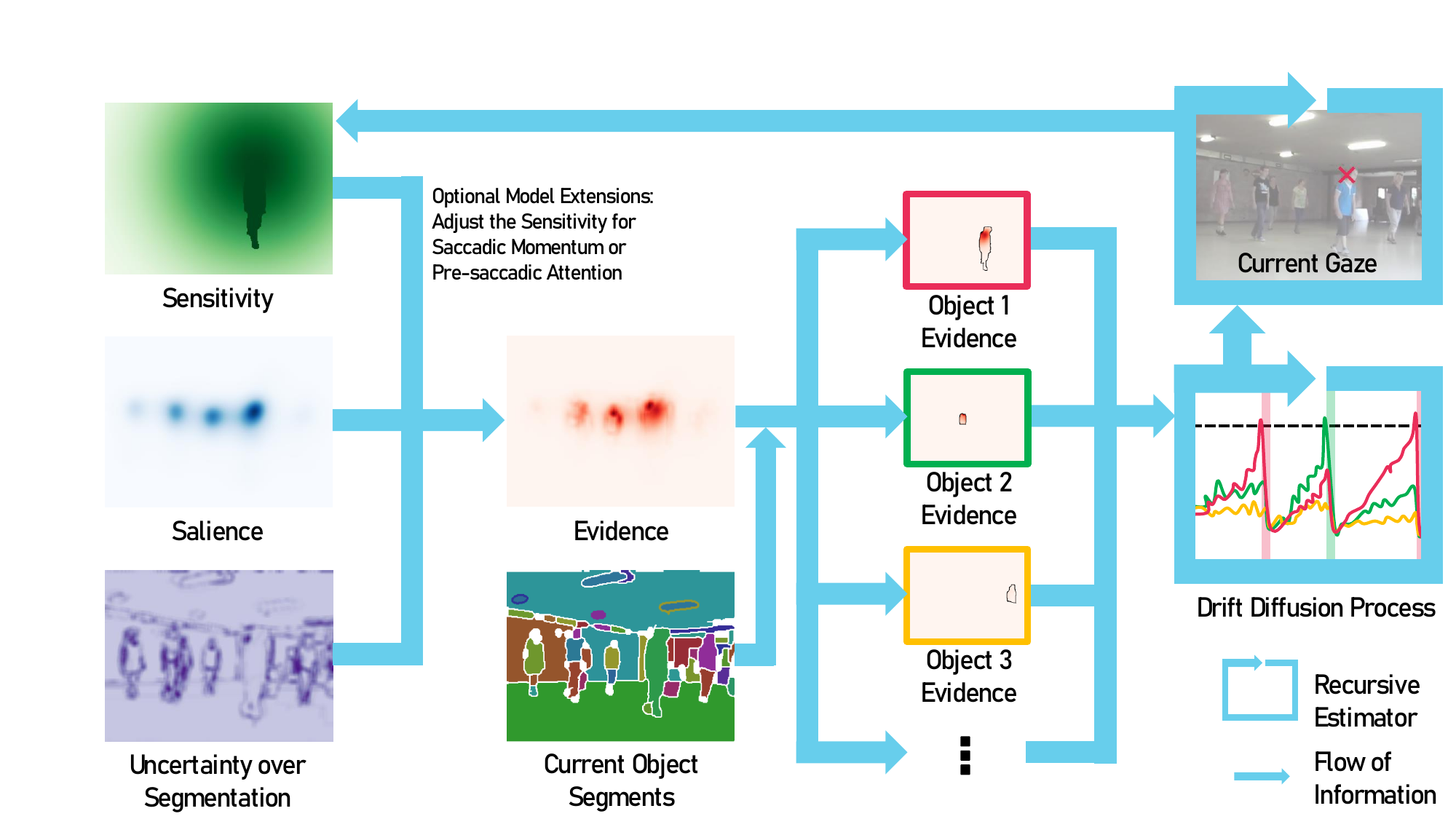}
    \caption{Our model makes saccadic decisions based on objects and is driven by uncertainty. It combines the uncertainty over object segmentation with salience and gaze-dependent sensitivity (left) into evidence for individual objects (middle). This evidence is then accumulated for each object in a drift-diffusion process (right). As soon as its threshold is passed, a saccade to this object is executed, otherwise the gaze smoothly pursues the currently foveated object. }
    \label{fig:scanpath_model}
\end{figure}

We model the saccadic decision-making process by adapting the object-based \emph{Scan}path simulation in \emph{Dy}namic scenes (\emph{ScanDy}) framework \citep{roth2023objects}.
The scanpath simulation updates its internal state, which includes a decision variable for all potential target objects in the scene, as segmented through the particle filter (\Cref{fig:object_seg_model}). 
We model the target selection process of where and when to move the gaze position with a drift-diffusion model (DDM), in which each object represents a potential saccade target (cf.~\Cref{fig:object_seg_model}).
The decision variable for each object depends on its eccentricity given the current gaze position, how relevant the visual scene features are, as measured by salience, and the uncertainty of the local object boundaries, as provided by the segmentation particle filter.  
Notably, the model does not rely on an explicit implementation of the ``inhibition of return'' (IOR) mechanism.

\paragraph{Scene relevance based on salient features}
We quantify the relevance of the scene content for gaze behavior by computing frame-wise feature maps $F$. 
Since we model free-viewing gaze behavior, where the observers have no explicit task, we approximate the relevance of different parts of the scene through visual saliency. 
We used the video saliency model UNISAL \citep{droste2020unified}, which was jointly trained on both image and video visual saliency datasets, since it is lightweight and produces state-of-the-art results on the DHF1K Benchmark \citep{wang2018revisiting}. We inferred the video saliency maps using the model with the domain adaptation for the DHF1K video dataset, which is most similar to the videos used in this study. The resulting video saliency predictions used as frame-wise feature maps $F(x,y)$ are normalized to $[0,1]$. $F$ is typically strongly localized around the most salient object (cf.\ \citealp{droste2020unified}). To allow the model to rely less on this strongly focused map, we introduce a model parameter $f_\mathrm{min}$ that linearly scales $F$ to $F'\in [f_\mathrm{min},1]$. By reducing the effective value range, a higher $f_\mathrm{min}$ parameter decreases the influence of the salience on the saccadic decision-making process.

\paragraph{Gaze-dependent visual sensitivity}
The foveation of the human visual system leads to a decrease in visual sensitivity with eccentricity from the current gaze position. As in \citet{roth2023objects}, we model the visual sensitivity $S$ across the scene with an isotropic Gaussian $G_\mathrm{S}$. We account for the well-documented object-based attentional benefit \citep{egly1994shifting, scholl2001objects, malcolm2015object} by approximating the covert spread of attention across the currently foveated object $O_\mathrm{f}$ (1 if pixel is part of the object, 0 if not) with a uniform sensitivity, replacing the part of $G_\mathrm{S}$ that falls within $O_\mathrm{f}$. 

\begin{align}
        G_\mathrm{S}(x,y) &= \frac{1}{2\pi\sigma_\mathrm{S}^2}\exp\left(-\frac{(x-x_0)^2+(y-y_0)^2}{2\sigma_\mathrm{S}^2}\right) \\
        S(x,y) &= 
        \begin{cases}
            1,& \text{if } O_\mathrm{f}(x,y)= 1\\
            G_\mathrm{S},& \text{else,}
        \end{cases}
    \end{align}
where the standard deviation $\sigma_\mathrm{S} = 7$ dva is set according to similar models (cf.\ \citealp{roth2023objects, schwetlick2020modeling}) and based on preliminary model explorations. 

In addition, we implemented two possible model extensions, which are not part of the base model but can be incorporated into the visual sensitivity $S$, namely \emph{saccadic momentum} and \emph{pre-saccadic attention}. 
In our explicit implementation of saccadic momentum, we increase the visual sensitivity in the direction of the previous saccade by generating an angle preference map based on the current gaze position and the angle of the previous saccade. We set a maximal value, which will be the sensitivity value in the direction of the previous saccade angle, that decreases linearly with the angle within a specified angle range to a minimum value. The resulting map (see \Cref{fig:extension_sensitivity}a) is then multiplied with $S$.
In our implementation of pre-saccadic attention, we assume a uniform spread of visual sensitivity across not only the currently foveated object but also objects that are likely to be the next saccade target (see \Cref{fig:extension_sensitivity}b). 

\paragraph{Uncertainty over object segmentation}
The visual system integrates different sources of information into a coherent visual representation of the environment \citep{milner2006visual}. If an object moves or input sources differ, for example, when the appearance and the motion-based segmentation find different object boundaries, this leads to a disagreement between instances in the segmentation particle filter. 
We include the resulting uncertainty over the object segmentation as the third contributing factor for the saccadic decision-making process, in addition to the relevance of the scene features and the gaze-dependent visual sensitivity. The uncertainty measure is directly obtained from the entropy $H(x,y)$ of the previously obtained boundary likelihood threshold in the object segmentation particle filter (see \Cref{eq:entropy}). We smooth the resulting map with a Gaussian blur, so uncertainties at the object boundaries are attributed to both objects. 
The values in the uncertainty map are, by construction, in the range $U(x,y) \in [0,1]$. Analogous to the scaling of the scene feature map, we introduce the model parameter $u_\mathrm{min}$ that linearly scales $U$ to $U'\in [u_\mathrm{min},1]$. Higher values for $u_\mathrm{min}$ hence effectively downscale the influence of the object uncertainty on $U'$.
The uncertainty at the current gaze position is typically low since the prompted segmentation of the currently foveated object provides a refined object mask, which is incorporated in the particle filter with high confidence. Through this interaction, the uncertainty contribution encourages the exploration of other objects in the scene.

\paragraph{Saccadic decision-making process}
We describe gaze behavior as a sequential decision-making process where objects in the scene accumulate evidence for becoming the next saccade target over time. As in the \emph{ScanDy} framework \citep{roth2023objects}, we model this latent cognitive process using a modified drift-diffusion model \citep[DDM;][]{ratcliff2016diffusion} with multiple options. The DDM accumulates evidence for each object over time (drift), while random fluctuations perturb each decision variable (diffusion). Unlike a classic DDM model, which includes only one decision variable and two thresholds for alternative choices, our model assigns each potential target object $i$ a decision variable $V_\mathrm{i}$ that accumulates toward a shared decision threshold $\theta$ (see \emph{Drift Diffusion Process} in \Cref{fig:scanpath_model}). As soon as the accumulated evidence for one object exceeds $\theta$, a saccade to this target is initiated. Hence, the DDM by design does not only model where but also when the eyes move.
The DDM drift rate $\mu_\mathrm{i}$ for an object at a given time depends on the task relevance based on scene features $F'(x,y)$, the visual sensitivity depending on the current gaze position $S(x,y)$, and the uncertainty of the object segmentation $U'(x,y)$. We multiply these maps in every frame to an evidence map $E(x,y,t) = S \cdot F' \cdot U'$, as shown in \Cref{fig:scanpath_model}. Next, we calculate $\mu_\mathrm{i}$ for each object mask $O_\mathrm{i}$ (1 if pixel is part of the object, 0 if not) in the resulting object segmentation of the particle filter (see \Cref{fig:object_seg_model}) as the average evidence across the mask $\Bar{E}(O_\mathrm{i}, t)$, scaled by the area $A_\mathrm{i}$ of the object, with

\begin{align}
    \Bar{E}(O_\mathrm{i}, t) &= \frac{ \sum_{x,y} E(x,y,t) } {\sum_{x,y} O_\mathrm{i}(x,y,t)}, \\
    A_\mathrm{i}(t) &= \sum_{x,y} O_\mathrm{i}(x,y,t) \cdot \nicefrac{1 \text{ dva}}{1 \text{ px}}^2, \\
    \mu_\mathrm{i}(t) &= \Bar{E}(O_\mathrm{i}, t) \cdot  \max\left(1, \log_2 A_\mathrm{i}(t)\right).
\end{align}
We convert the area from px$^2$ to dva$^2$ to ensure that videos with different resolutions are treated appropriately and scale the object's perceptual size logarithmically (cf.\ \citealt{nuthmann2017well}) to account for the difference in object sizes. 

The decision variable $V_\mathrm{i}$ for each object is then updated based on $\mu_\mathrm{i}$ and random fluctuations in the diffusion term $\epsilon \sim \mathcal{N}(0,\,1)$, with
\begin{equation}
	\label{eq:decision_var}
    V_\mathrm{i}(t+\Delta t) =  V_\mathrm{i}(t) + \nu \cdot ( \mu_\mathrm{i}(t) \Delta t + s \epsilon \sqrt{\Delta t} ),
\end{equation}
where the noise level $s$ is a free parameter, and $\nu$ is the fraction of time within $\Delta t$ spent on foveation since no evidence is accumulated during saccades. We set the update time resolution $\Delta t = 1$, measured in frames. We assume a linear update in $V_\mathrm{i}$ and can hence calculate the exact time when the decision threshold $\theta$ is crossed. As soon as $\theta$ is reached, we reset all decision variables $V_\mathrm{i}=0 \ \forall i$, and a saccade is executed to the corresponding object.
The saccade duration $\tau_\mathrm{s}$ scales linearly with the saccade amplitude $a_\mathrm{s}$ \citep{collewijn1988binocular, roth2023objects} with 
\begin{equation}
    \tau_\mathrm{s} = 2.7 \ \nicefrac{\text{ms}}{\text{dva}} \ \cdot a_\mathrm{s}  + 23 \text{ ms}.
\end{equation}

\paragraph{Gaze update}
We update the simulated gaze position at each time step (i.e., video frame). If the DDM threshold $\theta$ is not reached, the gaze point moves with the optic flow at its current position. This results, depending on the object and camera motion in the video, in either fixation or smooth pursuit behavior where the gaze moves with the object. 
If an $i$ exists with $V_\mathrm{i} > \theta$, a saccade is triggered to $O_\mathrm{i}$. The exact landing position within $O_\mathrm{i}$ is determined probabilistically, with the probability $p_\mathrm{i}(x,y)$ of each pixel being proportional to the scene features $F$ and gaze-dependent visual sensitivity $S$: 
\begin{equation}
	\label{eq:landing_pos}
    p_\mathrm{i}(x,y) \sim O_\mathrm{i}(x,y,t_0) \cdot F'(x,y,t_0) \cdot S(x,y,t_0).
\end{equation}

\subsection{Dataset}
\label{sec:mandm_dataset}
 
We compared the simulated scanpaths with human eye-tracking data recorded under free-viewing conditions on videos of natural scenes.
We collected eye-tracking data from 10 participants (8 female; mean age: $34.4$ years, range: $23-69$) on 43 video clips from the Unidentified Video Objects (UVO, \citealp{wang2021unidentified}) dataset (10 used for parameter tuning, 33 used for testing the model; randomly split). The videos were selected to show diverse content and to have temporally consistent, densely annotated object masks for the first 90 frames (cf.\ \citealp{wang2021unidentified}).

We recorded eye-tracking data for the here-used videos with an Eyelink 1000+ tabletop system (SR Research, Osgoode, ON, Canada) with a 1000 Hz sampling rate, as part of an ongoing collaborative large-scale eye movement database (publication of full dataset in preparation). We presented the videos in a dark room on a wall-mounted 16:9 video-projection screen (size: $150\times 84$ cm, Stewart Luxus Series 'GrayHawk G4', Stewart Filmscreen, Torrance, CA) at a distance of 180 cm from the study participants. We used a PROPixx projector (Vpixx Technologies, Saint-Bruno, QC, Canada) operating with $1920\times 1080$ pixels resolution on its native vertical refresh rate of 120 Hz. All videos were shown with a 30 fps framerate and (depending on their aspect ratio) scaled to a size of maximally 38.2 dva horizontally or 21.5 dva vertically ($1536\times 864$ pixels) to avoid high eccentricities. Participants started each trial with a fixational control (red dot on a black background) at a random location within the area where the scene was shown. The video was presented as soon as the participant fixated the target location (tolerance radius of 2 dva), ensuring high data quality and variation in the initial gaze position. All participants provided informed consent according to the \cite{helsinki} prior to data collection.

\paragraph{Event detection algorithm}
Identifying saccades in gaze data in dynamic scenes with object and camera movement in the scene can be a challenging task due to the presence of smooth pursuit eye movements. Potentially large pursuit velocities lead to a high number of false positive saccade detections in classic velocity-based algorithms such as the Engbert-Mergenthaler (EM) algorithm \citep{engbert2006microsaccades}. 
We, therefore, used the state-of-the-art U'n'Eye neural network architecture \citep{bellet2019human} and fine-tuned the network to our dataset. We labeled saccades, foveations (combining fixation and smooth pursuit events), and post-saccadic oscillation (PSO) events for one randomly selected second per video from different subjects.
Detecting PSOs is important to reliably define the endpoint of a saccade and hence precisely determine the duration of a foveation event \citep{schweitzer2022definition}.
The U'n'Eye network, with the training data we provided, was not able to reliably detect PSOs. Hence, we used the PSO detection based on saccade direction inversion, as described by \citet{schweitzer2022definition}. This algorithm expects saccades in the format provided by the EM algorithm. We, therefore, ran both the EM and U'n'Eye saccade detection algorithms, determined the saccades that were detected with both algorithms and then specified the exact saccade endpoint using the direction inversion criterion for PSO detection. 

\subsection{Metrics and parameter fitting}\label{sec:mandm_fitting}
 
We determined the model parameters by comparing the foveation duration and saccade amplitude distributions of the simulated scanpath with the human ground truth. 
We measured the similarity between a simulated distribution $N$ to the ground truth $M$ using the two-sample Kolmogorov-Smirnov (KS) statistic $D = \sup_x|N(x) - M(x)|$. We systematically varied the DDM noise level $s$, the decision threshold $\theta$, and the relative importance of the feature map $F'$ and uncertainty map $U'$, quantified by the re-scaling parameters $f_\mathrm{min}$ and $u_\mathrm{min}$. We performed a coarse grid search in this four-dimensional parameter space on the 10 videos in the training set. We simulated five different scanpaths (different random seeds) for each parameter configuration and compared them. Since we were particularly interested in the effect of uncertainty on the simulated scanpaths, we refined the grid search for each $u_\mathrm{min}$ value around the parameter sets leading to the lowest mean of the KS statistics for the foveation duration $D_\mathrm{FD}$ and the saccade amplitude $D_\mathrm{SA}$. We present an overview of all fixed and fitted model parameters, the parameter grid, and details of the fitting procedure in \ref{app:parameter_exploration}.

With the model parameters chosen such that the basic scanpath summary statistics of foveation durations and saccade amplitudes matched the human data, we evaluated the simulated scanpaths out-of-domain on the test set, i.e., on 33 previously unseen videos and on different metrics than what the parameters on the training set were selected for. For each parameter set, we simulated 10 scanpaths and compared them with the data from the 10 human observers.
We focused on the analysis of how gaze behavior balances the exploration of the background of a scene (\emph{Background}), uncover an object for the first time for foveal processing (\emph{Detection}), explore further details of the currently foveated object by making a within-object saccade (\emph{Inspection}), or return to a previously uncovered object (\emph{Return}) \citep{linka2021detection, roth2023objects}. Comparing the foveation durations in each category provides an insightful metric of the exploration behavior, which is particularly suited for dynamic scenes \cite[see][]{roth2023objects}. 
In addition to evaluating models on the test set, we also chose the later described \emph{base model} among different uncertainty values $u_\mathrm{min}$ based on this metric on the training set (see \Cref{sec:results_uncertainty} and \ref{app:parameter_exploration} for more details).

Since our model does not have an explicit IOR mechanism, we were particularly interested in whether it could reproduce typical IOR effects. IOR describes the inhibition of recently attended stimuli and the resulting delayed response to them \citep{posner1984components, klein2000inhibition}. In a free-viewing condition, as in the data used for this study, we therefore expect that saccades that return to a previous gaze position require more time to prepare. Hence, we analyzed the distribution of relative saccade angles of the human and simulated scanpaths. We divided all foveation events into 30 bins depending on the relative angle of the previous saccade (i.e., bin size of $12^{\circ}$). With the expectation that foveations preceding a return saccade ($\pm 180^{\circ}$) would be longer and foveations preceding forward saccades ($\pm 0^{\circ}$) would be shorter, we calculated the median foveation duration for each relative saccade angle bin. We used the median foveation duration instead of the mean to avoid a few very long events (in particular, smooth pursuit events can last multiple seconds), distorting the statistics for individual bins.

\section{Results}\label{sec:results}
 
Our aim was to build an image-computable and mechanistic computational model that closely resembles human gaze behavior in dynamic real-world scenes. In this section, we compare our model with human scanpaths, first qualitatively in \Cref{sec:results_qualitatively} and then quantitatively by reviewing aggregated statistics in \Cref{sec:results_aggregated}. We systematically explore the influence of uncertainty on visual exploration behavior in \Cref{sec:results_uncertainty}. 
In an additional ablation study, we probed the impact of individual object representations as input sources and the importance of the interaction between object perception and saccadic decision-making for the simulated scanpaths, as described in \Cref{sec:results_objects}. 
Lastly, we show how our model can be easily extended with additional modules, such as saccadic momentum or pre-saccadic attention, leading to more human-like saccade angle statistics and slight improvements in early object detections (\Cref{sec:results_extensions}).

\subsection{Qualitative scanpath analysis}\label{sec:results_qualitatively}

\begin{figure}
    \centering
    \includegraphics[width=1\linewidth]{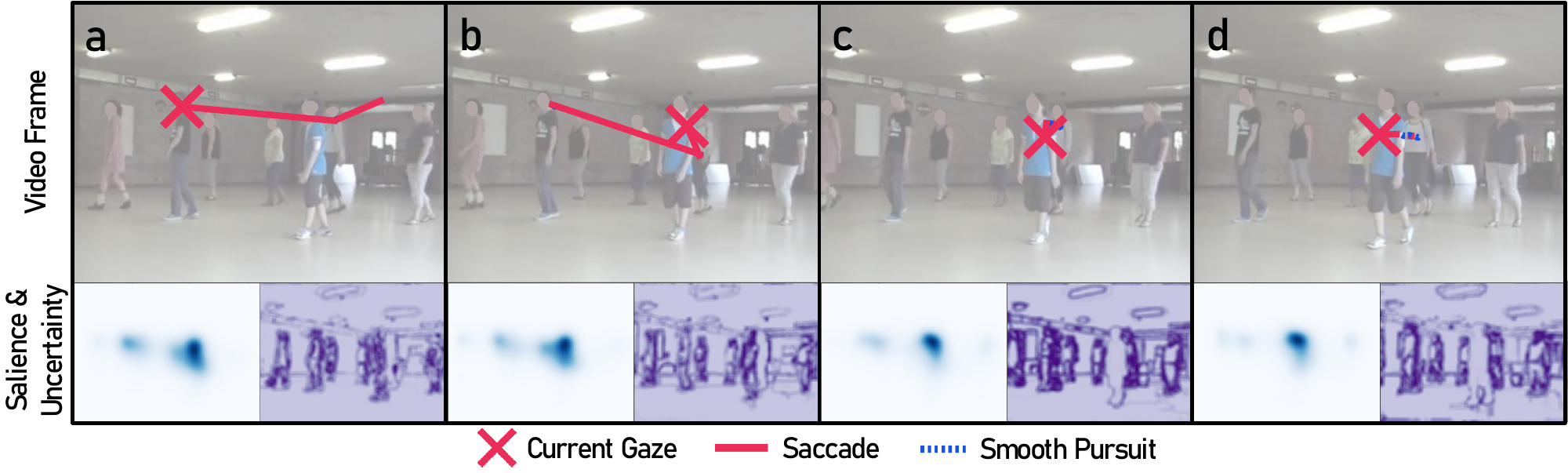}
    \caption{The predicted scanpaths of our model show human-like exploration in dynamic scenes. In this video of the test dataset, the model first follows uncertainty and detects two novel objects (dancers) (a), then returns to the first before detecting another one (b), which is then further inspected primarily due to its high visual saliency
    (c and d). For a video version, see \ref{app:videos}.}
    \label{fig:qualitative}
\end{figure}

Our model produces scanpaths that qualitatively closely resemble human visual exploration behavior; one example scanpath is shown in \Cref{fig:qualitative} (see videos in \ref{app:videos} for more examples and direct comparisons with human scanpaths). The access to individual mechanistic components of our model makes the individual saccadic decisions transparent and interpretable: Initially, all unexplored salient objects have relatively high uncertainty, which is resolved through large saccades towards them (\Cref{fig:qualitative}a; for a more detailed analysis of the uncertainty development, see \ref{app:uncertainty}).
Objects with particularly high saliency are likely to be revisited (\Cref{fig:qualitative}b) or are further inspected  (\Cref{fig:qualitative}c-d). 
Return saccades to previously foveated objects also become more likely with time, as uncertainty over object boundaries can rise again, for example, through object motion. This qualitatively similar behavior of our model can also be seen in \ref{app:videos}, where we show the exact scanpath and all intermediate computational steps as videos for $10$ simulations of our model as well as for $10$ human participants. We now further quantify these qualitative similarities between the human and modeled gaze behavior by comparing summary statistics of human scanpaths with the model predictions across the whole dataset.

\subsection{Aggregated scanpath statistics}\label{sec:results_aggregated}
\begin{figure}[t]
    \centering
    \includegraphics[width=1\linewidth]{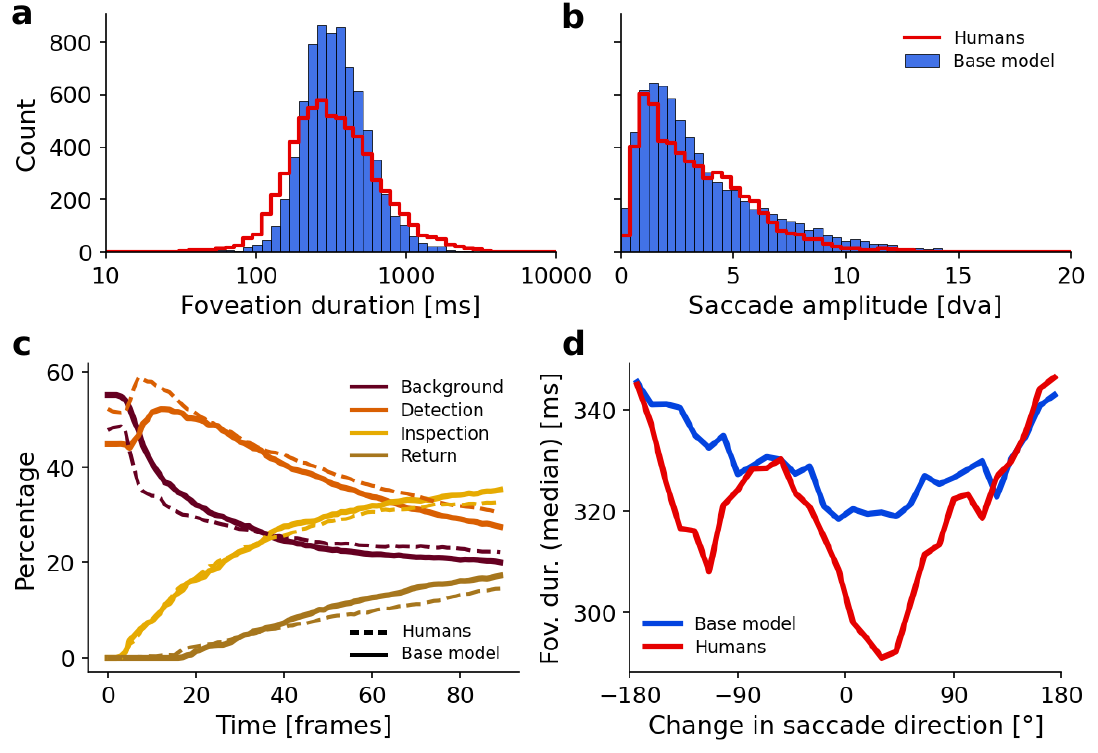}
    \caption{Aggregated statistics of the simulated scanpaths of the base model resemble the human eye-tracking data. (a) Histogram of the duration of all foveations in the human ground truth data (red) and the base model (blue). 
    (b) Histogram of the saccade amplitude distributions. 
    (c) Percentage of foveation events in the categories “Background” (maroon), “Detection” (orange), “Inspection” (yellow), and “Return” (khaki) across all human (solid) and model (dashed) scanpaths as a function of time. 
    (d) Median duration of the preceding foveation durations for each saccade. We applied a centered circular moving average across 5 bins (12$^{\circ}$ bin size) to reduce fluctuations in the median.}
    \label{fig:base_stats}
\end{figure}

We compare our base model predictions to human scanpaths on a set of videos not used for parameter search. As described before, we selected the model parameters to resemble the statistics of human foveation duration and saccade amplitude on 10 videos. The model generalizes well to the previously unseen set of 33 videos, as shown based on the aggregated scanpath statistics in \Cref{fig:base_stats}.
Similar to human eye-tracking data, the foveation durations (\Cref{fig:base_stats}a) of the simulated scanpaths follow a log-normal distribution with a mean of $390$ ms and a median of $332$ ms (humans: mean of $433$ ms and median of $316$ ms). The distribution of the model is more narrow compared to humans, which---if other metrics would not be considered---could be corrected by increasing both the decision threshold and the noise level in the drift-diffusion model, as described in \Cref{sec:mandm_scan}. 
The saccade amplitudes in the simulated scanpaths (\Cref{fig:base_stats}b) follow the gamma distribution of the human data with a mean of $3.70$ dva and a median of $2.81$ dva (humans: mean of $3.40$ dva and median of $2.90$ dva). These well-described statistics are not explicitly implemented in the model, but emerge from model constraints: Foveation durations are a consequence of the way evidence is accumulated in the decision-making process. Saccade amplitudes result from the balance of local exploration, as encouraged by the visual sensitivity, and global exploration, as driven by uncertainty and noise in the drift-diffusion model (DDM). 

Besides replicating these basic summary statistics, we are interested in how the exploration behavior of our model compares to that of humans. 
Our model, like the participants in our dataset, starts at a random initial location on the scene. Hence, about half of the scanpaths start on the background. In \Cref{fig:base_stats}c, we can observe that the model---similar to humans---quickly starts to favor the exploration of novel objects (detections) rather than further exploring the background. We additionally confirmed that in a large proportion of the scenes, the object that was first detected by the majority of human observers was also first detected in the majority of simulated scanpaths (24 of 33 scenes in the test set, $72.7\%$ agreement; base rate: $23.6\%$, estimated as the average of $\frac{1}{N_{0,s}}$ across scenes, where $N_{0,s}$ is the number of objects in the first frame of each scene).
After an initial peak in detections in both models and humans, the amount of detection decreases in both cases in favor of further saccades within the currently foveated object (inspections) or revisits of previously foveated objects (returns).
Overall, both in relative amounts and trends over time, the balance between the exploration of the background, new objects, and already-seen objects of the model resembles the human behavior well. Typically, such a balance in exploration can be achieved through a suitable parametrization of an explicitly implemented ``inhibition of return'' (IOR) mechanism (cf., \citealp{itti2001computational, roth2023objects}). We find that in our model, the relative influence of the uncertainty map plays a crucial role in achieving this balance, which we describe in detail in \Cref{sec:results_uncertainty}.

The model even shows the expected \emph{temporal} IOR effect \citep{klein1999inhibition, luke2014dissociating}, as shown in \Cref{fig:base_stats}d, without explicitly implementing it and without adjusting any parameters to reproduce this statistic. We find a characteristic dip in foveation durations before a saccade is executed in the same direction as the previous saccade (forward saccades), as observed in the human scanpaths. The preparation of saccades with larger turning angles is slower.
This is a result of the uncertainty at the previous gaze position being reduced through the foveated object cue, such that the accumulation of evidence will take longer for a return saccade (for a more detailed analysis, see \Cref{sec:results_uncertainty}).

In summary, our model also quantitatively resembles human scanpaths in dynamic scenes, both in its basic statistics and its exploration behavior. In the next two sections, we further show that the similarities between human and modeled scanpaths, particularly the exploration behavior balances between the different functions of foveations, can be attributed to two features of our model: The consideration of uncertainty and bidirectional interaction between object perception, and saccadic decision-making to generate appropriate perceptual units to operate on.

\subsection{Model ablation 1: Uncertainty drives exploration}\label{sec:results_uncertainty}

\begin{figure}[!ht]
    \centering
    \includegraphics[width=1\linewidth]{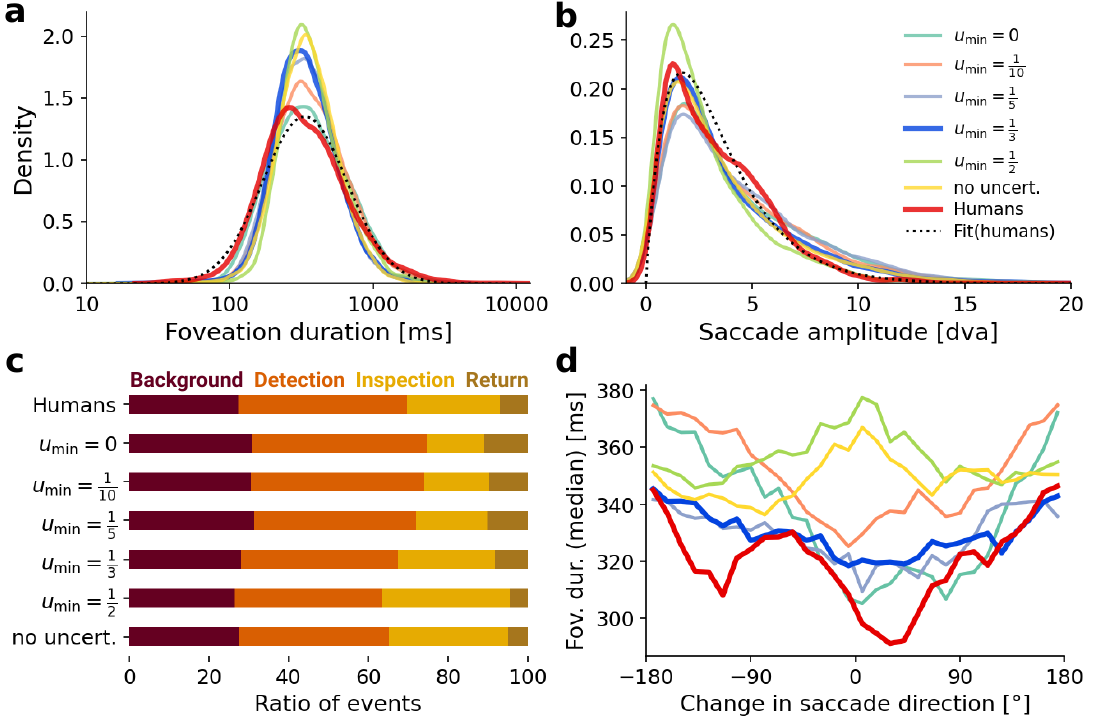}
    \caption{The uncertainty contribution in the model determines the exploration behavior. 
    (a) Kernel density estimation (KDE) of the distribution of foveation durations for the human data and simulated scanpaths with different uncertainty contributions. The dotted line indicates a log-normal fit to the human data with $\mu=5.815$ and $\sigma=0.681$ (equiv.~to an expected value of $e^{\mu+\frac{\sigma^2}{2}}=422.8$ ms). (b) KDE for the saccade amplitude distributions with a fitted Gamma distribution to the human data with shape $\alpha=2.01$ and rate $\beta=0.59$ (equiv.~to an expected value of $\frac{\alpha}{\beta}=3.40$ dva). (c) Ratio of time spent in the different foveation categories, as shown in \Cref{fig:base_stats}c, averaged across time. (d) Temporal IOR effect for the different uncertainty contributions, as in \Cref{fig:base_stats}d.
    The model with $u_\mathrm{min}=\frac{1}{3}$ corresponds to the base model in \Cref{fig:base_stats}. Further information about the individual model parameters can be found in \ref{app:parameter_exploration}.}
    \label{fig:uncert_stats}
\end{figure}

Our model uses the uncertainty measure of the object segmentation module as an estimate for the perceptual uncertainty that influences the gaze behavior depending on the scanpath history.
Here, we evaluate the effect of this uncertainty mechanism, comparing the simulated model scanpaths with a varying influence of the uncertainty on the saccadic decision-making process. Specifically, we vary the $u_\mathrm{min}$ parameter of the model where a \emph{higher} value \emph{decreases} the importance of uncertainty. We also compare results from this model with those of a version that does not consider uncertainty at all.

We select the threshold $\theta$ and the noise level $s$ of the DDM for each value of $u_\mathrm{min}$ anew to fit the foveation duration and saccadic amplitude statistics (see \Cref{sec:mandm_fitting}). Hence, varying the importance of uncertainty does not strongly influence the basic scanpath statistics, as shown in \Cref{fig:uncert_stats}a-b. While the specific densities change, the general shape of the log-normal (foveation duration) and gamma (saccade amplitude) distributions remain stable. However, how the model balances exploration behavior changes considerably (\Cref{fig:uncert_stats}c): For high importance of uncertainty ($u_\mathrm{min}<\frac{1}{3}$), the model focuses on the exploration of previously unvisited parts of the scene (background, detection) or returns to previously detected objects while only rarely further inspecting the currently attended object. For low importance ($u_\mathrm{min}>\frac{1}{3}$ or ``no uncert.''), on the other hand, we observe the opposite trend, where the currently attended object is inspected further because the uncertainty over other parts of the scene does not drive the exploration there. For this reason, we find the right trade-off between the exploration of novel parts and the further inspection of attended parts at a medium importance of $u_\mathrm{min}=\frac{1}{3}$.

If we further analyze the temporal IOR effect under the different variations of uncertainty importance (\Cref{fig:uncert_stats}d), we observe that the model requires a certain level of uncertainty importance (at least $u_\mathrm{min}=\frac{1}{3}$) to reach the same effect. With too low importance of uncertainty ($u_\mathrm{min}>\frac{1}{3}$ or ``no uncert.''), the effect is instead inverted, such that saccades in the same direction are preceded by longer foveation durations. Hence, incorporating uncertainty into the model increases the probability of return events while simultaneously giving rise to the temporal IOR effect.
The capability of accounting for this effect highlights how uncertainty can replace an IOR mechanism that is explicitly built into the model to drive human-like exploration behavior. The uncertainty of our model is computed based on the different object cues the model receives to build the object segmentation of the scene. In the following, we analyze the influence of these different inputs and how the resulting object representations affect the simulated scanpaths.

\subsection{Model ablation 2: Semantic object cues and component interconnections form suitable perceptual units}\label{sec:results_objects}

\begin{figure}[t]
    \centering
    \includegraphics[width=1\linewidth]{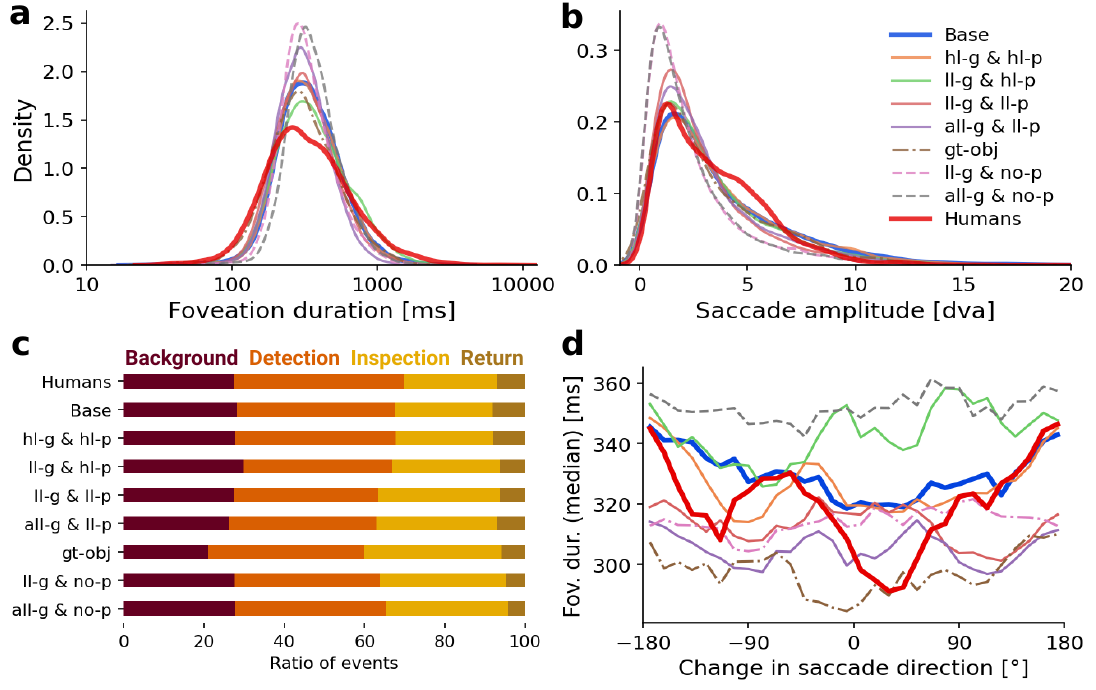}
    \caption{Semantic object cues and the interconnection through the gaze-dependent prompt are crucial for human-like simulated scanpaths.
    (a-d) Analogous to \Cref{fig:uncert_stats} for models that use different object cues in the segmentation module. We compared the human data and the base model with models that use only the high-level/semantic object cues for the global and the prompted segmentation (\emph{hl-g \& hl-p}), only the low-level/appearance-\& motion-based global segmentation and the high-level prompt (\emph{ll-g \& hl-p}), a low-level/appearance-based prompt either combined with only low-level or with all global cues (\emph{ll-g \& ll-p}, \emph{all-g \& ll-p}), a model that uses ground truth objects together with the base model uncertainty (\emph{gt-obj}), and models that use either only low-level or all global object cues without any prompted object (\emph{ll-g \& no-p}, \emph{all-g \& no-p}).}
    \label{fig:object_ablation}
\end{figure}

Our model updates both the current object segmentation and perceptual uncertainty from the current image of the scene using different object cues. The segmentation then defines the perceptual units in saccadic decision-making, and the uncertainty influences the likelihood of selecting these perceptual units. In this ablation, we investigated the extent to which different object cues in the segmentation algorithm affect the predicted scanpaths. 
We compared different combinations of low-level (basic appearance and motion) and high-level cues (with semantic/top-down influence) for both the global and gaze-dependent object segmentation. 
The primary scanpath statistics did not change by much if we replaced the object sources (\Cref{fig:object_ablation}a-b). For simplicity (high computational cost of the parameter exploration) and since we are primarily interested in the overall trends, we used the same model parameters as in the base model. Only for the models without a prompted object segmentation (i.e., in which uncertainty is not lowered at the current gaze position), we found a new parameter set for the comparison (see \ref{app:parameter_exploration} for more details).
We compared the resulting scanpaths to our base model (low- and high-level global segmentation combined with a high-level prompted mask, i.e., \emph{all-g \& hl-p}), a model that uses a ground truth object segmentation (provided as labels in the dataset, cf.~\Cref{sec:mandm_dataset}), and the human scanpaths.

We found that a model that does not use any low-level object cues but instead relies only on the high-level, semantic global segmentation and the semantic prompt (\emph{hl-g \& hl-p}) explored the scene very similarly to the base model and the human observers in terms of both the functional categories (\Cref{fig:object_ablation}c) and the temporal IOR effect (\Cref{fig:object_ablation}d). 
If, instead, only appearance- and motion-based segmentations were used as global object cues (\emph{ll-g \& hl-p}), the exploration behavior of the model remained close to the human data as long as foveated segmentations take advantage of high-level cues. The lack of a global semantic segmentation, however, led to more exploration of the background due to the uncertain low-level segmentation and, thus, made the characteristic dip in the temporal IOR effect disappear.
We also implemented a model that used exclusively low-level object cues by replacing the high-confidence prompted object segmentation with an appearance-based low-level object prompt (\emph{ll-g \& ll-p}). In this case, the model segmented individual pieces of clothing based on color when foveating a person. If, instead, semantics were used, the person, including their clothes, would be considered a single object. This, in turn, would lead to a higher number of inspections as there is more uncertainty within the remaining ground truth objects.
Adding the global semantic segmentation to the model with low-level prompts had almost no effect on the scanpath statistics (\emph{all-g \& ll-p}).

We next investigated the influence of the interaction between saccadic decisions and segmentation. We removed one of the two directions of these interactions. First, we replaced the perceptual units generated by our segmentation component with the ground truth objects provided by the dataset, while still computing and using the uncertainty map as in the base model. 
As a result, the few labeled objects were often and reliably foveated, leading to a high amount of inspections while the background was explored much less. 
In an additional ablation, we removed the foveated segmentations from our model (\emph{all-g \& no-p} and \emph{ll-g \& no-p}), using the particle filter for the global segmentations but making the segmentation into perceptual units independent of the gaze. Hence, we removed the ability of the model to actively resolve uncertainty through saccades. This changed its exploration behavior considerably: Inspections became much more frequent, while detection times decreased. Moreover, we no longer observed any temporal IOR effect.

In summary, we found that removing low-level object cues from the segmentation filter does not lead to big changes in the resulting scanpaths. High-level semantic segmentation cues, however, were needed to simulate human-like gaze behavior. In particular, high-level prompted object cues entailed a temporal IOR effect.
When we removed the ability of the model to reduce the object uncertainty through saccadic decisions through the prompted object cue, we observed an even larger effect on the simulated scanpaths. Even if the model included a global semantic segmentation, the uncertainty-driven interaction between the two components was crucial.

\subsection{Model extensions: Saccadic momentum improves saccade angle statistic and pre-saccadic attention benefits early object detections}
\label{sec:results_extensions}

\begin{figure}[t]
    \centering    \includegraphics[width=1\linewidth]{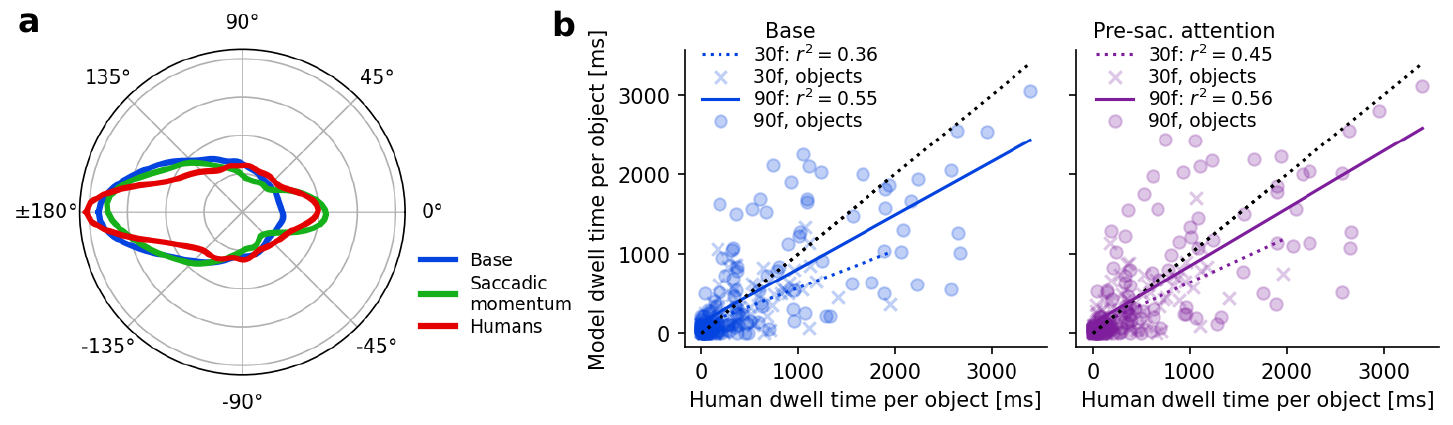}
    \caption{Extending the model through saccadic momentum or pre-saccadic attention leads to improvements in certain statistics. 
    (a) Histograms of the change in saccade direction for scanpaths simulated with the base model (blue) and the model with saccadic momentum (green) compared to the human data (red). Forward saccades with $\pm 0^{\circ}$ go in the same direction, while return saccades with $\pm 180^{\circ}$ go in the opposite direction compared to the previous saccade.
    (b) Dwell time for each individual object averaged across human observers compared to simulated model scanpaths of the base model (left, blue) or the model with pre-saccadic attention (right, purple). We distinguish between the time objects were foveated in the first 30 frames (marked with \emph{x}, dotted regression line) and in the first 90 frames (maximum number of frames with objects; marked with \emph{o}, solid regression line). A perfect prediction would correspond to the data points for all objects lying on the dotted line with slope $m = 1$ and intercept $y_0 = 0$.
    See \Cref{fig:extension_stats} in \ref{app:extensions} for the aggregated scanpath statistics analogous to Figs.~\ref{fig:uncert_stats} and~\ref{fig:object_ablation} of the extended models.}
    \label{fig:extension_showcase}
\end{figure}

We have shown so far that our model reproduces important hallmarks of scanpaths in dynamic real-world scenes. One instructive metric we have not yet investigated is the distribution of relative saccade angles. Importantly, this distribution shows how many forward and return saccades were made and is therefore also interesting in the context of \emph{spatial} IOR, that is, the reduced probability of returning to a previously visited location. 
The human scanpaths in our data show a strong bias for saccades in the opposite direction relative to the previous saccades, as shown in \Cref{fig:extension_showcase}a. This is in line with work that showed that return saccades are much more frequent in complex scenes than expected from the IOR literature \citep{smith2009facilitation, burlingham2023motor}. 
Since our model does not explicitly inhibit return saccades, this behavior is replicated well. Yet, the base model did not reproduce the human bias to make saccades in the same direction as the previous saccades,  called \emph{saccadic momentum} \citep{anderson2008directional, smith2009facilitation}. Different mechanisms have been discussed to explain saccadic momentum, 
including a continuation of the motor plan and a visual bias in V4 neurons \citep{motter2018saccadic}.
While no such mechanism is implemented in the base model, its modular implementation makes it easy to account for the saccadic momentum effect.

We thus extended our base model by introducing a bias towards forward saccades into the gaze-dependent visual sensitivity $S$ (see \Cref{sec:mandm_scan}), while keeping all other model parameters the same. Unsurprisingly, the model with saccadic momentum reproduced the relative saccade angle distribution(\Cref{fig:extension_showcase}a). Importantly, the previously investigated statistics of human exploration behavior remained largely unaffected (see \Cref{fig:extension_stats}).

In a second extension, we included the well-established finding of \emph{pre-saccadic attention} shifts \citep{deubel1996saccade, rolfs2011predictive} into the model. We implemented this by prompting objects whose evidence exceeded 30\% of the DDM threshold $\theta$ and setting the sensitivity map $S$ for these objects to $1$, just as if they were foveated (see \Cref{fig:extension_sensitivity}b). Again, we added this component to the base model while keeping all other model parameters unchanged. Effectively, this provides the model with additional saliency information at the most likely saccade targets, which should help to better prioritize between them. Therefore, we expected this pre-saccadic attention model to be more consistent in exploring the same objects as the human observers than the base model. We did not see a considerable change in the correlation of the overall object-specific dwell time when considering the whole duration for which object masks are available (90 frames; $m=0.67, y_0=139.6, r^2=0.55$ for the base and $m=0.72, y_0=126.6, r^2=0.56$ for the pre-saccadic attention model; \Cref{fig:extension_showcase}). In an exploratory analysis where we only considered the objects foveated in the first second, which in human scanpaths primarily corresponds to detections of the most salient objects \citep[cf.][]{parkhurst2002modeling, donk2008effects}, we did see an improvement in the correlation through this model extension (30 frames; $m=0.47, y_0=101.1, r^2=0.36$ for the base and $m=0.56, y_0=76.9, r^2=0.45$ for the pre-saccadic attention model). We predict that this attentional benefit would become more pronounced if we were to fit all free model parameters again for the pre-saccadic attention model and/or specifically fit the models to reproduce the object-specific dwell times.

Finally, we accounted for the simplified assumption that a saccade is executed immediately after the decision threshold is reached. It has been shown that new visual information does not influence the movement plan anymore in the final 50-70 ms of the preceding fixation \citep{hooge1999peripheral, ludwig2005remote}. We implemented such a saccadic dead time in our model by prolonging each foveation by 50 ms, during which no evidence is accumulated. After this dead time, the saccade is executed as in the base model. Without fitting the parameters again, we only lowered the decision threshold $\theta=4.0$ (base model) to $\theta'=3.5$ to account for the otherwise 50 ms longer foveation durations, and keep all other parameters as in the base model. We find that the inclusion of this dead time does not make a qualitative difference on any of the investigated metrics (see \Cref{fig:extension_stats}).


\section{Discussion}
\label{sec:discussion}

We presented a model for object-based attention and gaze behavior in complex dynamic scenes that builds on a previous model for saccadic decision-making \citep{roth2023objects} and an object segmentation model for interactive perception in robotics \citep{mengers2023combining}. The active interconnection between the two model components resembles an algorithmic information processing pattern from robotics, Active InterCONnect (AICON; see \citealp{battaje2024aicon}), which we further examine in \Cref{sec:dis_aicon}. Prior to this, we discuss the results of our study (\Cref{sec:dis_summaryeval}) as well as the limitations and advantages of our approach (\Cref{sec:dis_study}). In particular, we elaborate on the conclusions we can draw about uncertainty as a driving factor for visual exploration (\Cref{sec:dis_uncertainty}), and what we can learn from the model about the perceptual units of visual attention (\Cref{sec:dis_units}).

\subsection{Summary and evaluation of the results}\label{sec:dis_summaryeval}

Our scanpath model successfully replicates key aspects of human visual exploration in dynamic real-world scenes. Qualitative (\Cref{sec:results_qualitatively}) and quantitative (\Cref{sec:results_aggregated}) comparisons between simulated and human gaze behavior demonstrate that the model closely resembles human behavior and accurately reproduces scanpath statistics. We selected the model parameters such that the simulated scanpaths fit the foveation duration and saccade amplitude statistics of human eye-tracking data. Without further fitting, the model captures meaningful exploration patterns on unseen videos, including the temporal balance between detecting new objects, inspecting currently foveated objects, and returning to previously viewed areas. This balance is primarily driven by the influence of uncertainty on saccadic decisions, which also leads to a temporal IOR effect without the need for an explicit implementation of an IOR mechanism (see \Cref{sec:results_uncertainty}). We further investigated how different object sources, such as low-level and high-level cues, influence scanpaths and found that semantic object cues played a crucial role in obtaining human-like exploration (see \Cref{sec:results_objects}). Additionally, model extensions incorporating psychophysically uncovered mechanisms like saccadic momentum and pre-saccadic attention have the potential to further align the model's resemblance to human behavior in terms of saccade angle distributions and object dwell time (see \Cref{sec:results_extensions}).

Combined, the scanpath evaluation metrics in this work offer a comprehensive view of how well the model mimics human gaze behavior by assessing both temporal and spatial dynamics in visual exploration. Ideally, a single metric would capture all aspects of the simulated behavior, but currently, no established evaluation metric exists for scanpaths in dynamic scenes. For models with a readily computable sequential likelihood function, data assimilation has shown promise as an approach for both parameter fitting and model evaluation \citep{engbert2022data, schwetlick2020modeling, seelig2020bayesian, schutt2017likelihood}. Although it might be conceivable to approximate the spatiotemporal likelihood function for our model's scanpaths and update them frame-by-frame, this approach would be computationally infeasible. In addition to recomputing the likelihood for every frame, it is unclear how to extend the point processes used in the sequential likelihood approach to include smooth pursuit events (for a detailed discussion on additional scanpath evaluation metrics in dynamic scenes, see \citealp{roth2023objects}).

\subsection{Advantages and limitations of our model}\label{sec:dis_study}
The here presented model still has many of the simplifications of our previous framework for \emph{Scan}path simulation in \emph{Dy}namic scenes (\emph{ScanDy}) \citep{roth2023objects}. Importantly, we assume that attention spreads instantaneously and uniformly across objects and that saccades are always precisely executed without attempting to model the saccade programming and oculomotor control.
While we focus on scene segmentation and scanpath simulation in the current work, our modular implementation should make it easy to further extend the model in that direction.
The current extensions of saccadic momentum and pre-saccadic attention both only required the addition of a few lines of code. 

So far, we have only modeled scanpaths during free-viewing, that is, observers had no task instructions. In the future, we plan to apply the same modeling approach to simulate scanpaths in complex dynamic scenes during goal-directed tasks, such as visual search and scene memorization. We expect that additional top-down attentional control during these tasks can be incorporated into the modeling by adapting the feature map $F$ (see \Cref{sec:mandm_scan}, $F$ currently represents only visual saliency) and tuning the model parameters.
For example, we would anticipate that our model could already reasonably simulate scanpaths for scene memorization through a down-scaling of the importance of $F$ through $f_\mathrm{min}$ and visual search through the inclusion of a target similarity map in $F'$. In both cases, the threshold of the drift-diffusion model (DDM) $\theta$ should be lowered to account for typically shorter foveation durations under such task conditions \citep{rayner200935th}.

The important improvement over the existing \emph{ScanDy} framework is the active interconnection with object segmentation. Through this interaction, the model becomes image-computable, that is, we do not have to define what constitutes an object a priori, but the object representations change based on the scanpath. The implementation of the object segmentation as a recursive Bayesian filter leads to a serial dependence of the segmentation, using both prior and present object information to represent the scene \citep{fischer2014serial}. Furthermore, the segmentation module
automatically provides us with an uncertainty map, which depends on the prior and present gaze position. We show that through the automatic reduction of uncertainty as a consequence of saccadic decisions, this uncertainty map is well suited to drive saccadic exploration behavior during dynamic free-viewing scenes.

Importantly, when we say we have a mechanistic model, we refer to attentional mechanisms in the sense of algorithmic principles and do not make claims on the biological or implementational level \citep[cf.][]{marr1982vision}. Although there is evidence for Bayesian updating in the brain \citep{knill2004bayesian, ma2006bayesian}, even in the form of a neural particle filter \citep{kutschireiter2017nonlinear}, we want to argue more conceptually for principled ways of information processing, independently of neural implementation. For example, there is evidence of bidirectional information exchange between different components of perceptual processing, similar to the exchanges between our components for object segmentation and saccadic decision-making. Such exchanges have been observed not only between different hierarchical levels of processing \citep{ahissar2004reverse}, but also laterally between the processing of different cues \citep{livingstone1988segregation} or even between separate sensory modalities \citep{mcgurk1976hearing}. 

In our model, we recursively update the segmentation in the object component and the evidence in the saccadic decision component. Hence, the model makes use of the temporal consistency of the visual environment, which has also been observed and described in human behavior during visual search \citep[e.g.,][]{niemi1981foreperiod,kristjansson2010fortune} and object perception \citep{liu2008learning,blake1997spatial}. For this segmentation, we aimed to combine and compare object cues based on low-level appearance \citep{schyns1994blobs}, motion \citep{reppas1997representation}, and semantics \citep{neri2017object}, which have been shown to play a role in the human visual system. While the recursive Bayesian integration of these object cues is mechanistically plausible, the way our model computes these inputs is certainly different from how the visual system might infer them. The computer vision algorithms used to obtain these cues, as described in \Cref{sec:mandm_segmentation_cues}, and particularly the semantic segmentation, on which we provide further details in \ref{app:semanticsegmentation}, were not chosen based on their biological plausibility but rather for how well their results represent the respective object cues as uncovered in psychophysical experiments. Similarly, the prompted semantic segmentation of the currently foveated object does not use a more plausible foveated input frame, since this would be outside the training distribution of the algorithm. Instead, we use a higher resolution of the input frame compared to the pre-attentive global segmentations, prompt the model at the current gaze position (see \ref{app:semanticsegmentation} for details), and include the resulting mask with higher confidence into the particle filter. An additional foveal benefit plays a role in the subsequent saccadic decision-making process, where the combination of global scene features $F$, and the gaze-dependent visual sensitivity $S$ approximates the incoming information at any point in time. Our model is hence plausible on the level of attentional mechanisms and utilized object cues, but not on the level of how these are currently implemented.

The modular and mechanistic design of the model allows us to explore essential hypotheses about attention and gaze behavior in dynamic scenes---which can be challenging to test experimentally. By studying the model's behavior, we can generate hypotheses that can later be tested in eye-tracking experiments specifically designed for this purpose. The model offers complete control over its internal processes, allowing us to perform various ablation studies, including those on latent variables, which are usually difficult to assess in behavioral experiments. 
In the interpretation of our model ablation results, we assume that the other parts of our model are mechanistically similar to the human visual system. This allows us to deduce how the investigated mechanism (i.e., the inclusion of uncertainty for gaze guidance or the formation of perceptual units for object-based attention) best interacts with the other model components to produce human-like gaze behavior in dynamic scenes.

In our implementation of attentional mechanisms, we focused on what we consider the core components of the vast literature on attentional guidance. In theory, including other mechanisms may change the interplay between model components and, as a result, the interpretation of our ablations. In practice, however, we find that--- while our extensions of the model improve certain statistics of the simulated scanpaths---they do not qualitatively change the model's overall behavior. While this is not a guarantee that it will be the same for future model extensions, it increases our confidence in the robustness of our model and its predictive power for mechanisms of visual attention.
Therefore, we can develop hypotheses about the inner workings of the human visual system by systematically examining how our model produces certain behaviors. These hypotheses can then be tested in psychophysical experiments guided by the model. In the following sections, we discuss two insights from the model and how they may inspire psychophysical experiments.

\subsection{Uncertainty drives exploration}\label{sec:dis_uncertainty}
 
The connection between active exploration behavior and the reduction of perceived uncertainty of the environment is well established in the literature \citep{sullivan2012role, friston2012perceptions, renninger2007look}. 
\citet{gottlieb2013information} summarized that ``information-seeking obeys the imperative to reduce uncertainty and can be extrinsically or intrinsically motivated'' (p. 586) and that ``the key questions we have to address when studying exploration and information-seeking pertain to the ways in which observers handle their own epistemic states, and specifically, how observers estimate their own uncertainty and find strategies that reduce that uncertainty'' (p. 586). 
It is, however, not obvious how uncertainty should be measured and quantified in an image-computable model of visual attention. 

In this context, it is important to clarify again what we mean by uncertainty since the term can refer to many things. Our model specifically considers the uncertainty of the boundary between potential objects, both about their existence and exact location, but not about the object's identity or other possible features (for more details, see \ref{app:uncertainty}).
For example, if an object in the periphery moves, this typically would increase the uncertainty estimate in our model. One could argue that the additional motion cue should reduce the uncertainty about the shape of the object. Indeed, this intuition is reflected in our model since the input from the motion segmentation will clearly show the object. However, the overall uncertainty of the object might still increase because the exact position, shape, or state of the moving object might change, which would be reflected in conflicting object measurements from different sources or in a strong deviation from the prior belief. This prior belief is calculated as the segmentation of the previous frame, shifted by the optical flow. 

Our results show that including the uncertainty map of the object segmentation module as a driving factor in the saccadic decision-making process leads to human-like simulated scanpaths. The weight of the uncertainty map for the decision-making process, parameterized through $u_\mathrm{min}$, strongly influences the ratio between foveation categories, in particular, the frequency at which objects are inspected. The prompted high-confidence object segmentation typically leads to a low uncertainty at the current position, encouraging further exploration of the scene and more return events for a strong influence of uncertainty (low $u_\mathrm{min}$). If the influence is weak (high $u_\mathrm{min}$), the gaze-dependent spread of attention leads to a strong tendency to further inspect objects with high salience. Interestingly, the $u_\mathrm{min}$ parameter also influences the strength of the temporal IOR effect. Despite returns occurring more often with a lower $u_\mathrm{min}$, the uncertainty of recently foveated objects is typically reduced, thereby slowing down the evidence accumulation process. While IOR is generally conceived as a viewing bias that both reduces (spatial) and delays (temporal) return events, our uncertainty-guided model  captures not only the temporal IOR but also the spatial ``facilitation of return'' \citep{smith2009facilitation} observed in the human scanpaths.

Most mechanistic scanpath models require an explicit implementation of IOR (cf.\ \citealp{itti1998model, zelinsky2008theory, schwetlick2020modeling, roth2023objects}) to avoid being bound to the objects or locations with the highest salience \citep{itti2001computational}. 
Our model takes a different approach, similar to previous computational models that have incorporated uncertainty-based strategies, where exploration is driven by high variance or entropy \citep{cohn1996active, rothkopf2010credit}. It is closely related to the principle of information maximization, which has been applied before to simulate eye movements in static scenes \citep{renninger2004information, lee1999information, wang2011simulating}. Where our model is uncertain is also closely related to ``Bayesian surprise'', which was introduced by \cite{itti2009bayesian} in the context of scanpaths as a measure for how eye movement data affects differences between posterior and prior beliefs of an observer about the world.
These models also do not require an explicit IOR implementation, since there is little information to be gained by revisiting already foveated parts of the scene. However, when observing dynamic real-world scenes, further inspections and returns are frequent, and defining an information maximization or uncertainty-driven approach that can account for this behavior is not trivial.
In our model, we do not need a separate estimation of the uncertainty, since it is a natural by-product of the AICON-ic way in which we obtain the object segmentation.

 \subsection{Perceptual units for object-based attention}\label{sec:dis_units}

Object-based attention is a well-established concept that has been thoroughly investigated in a large variety of experimental paradigms \citep{scholl2001objects, peters2021capturing, cavanagh2023architecture}. However, it remains unclear what constitutes a visual object in this context \citep{spelke1990principles, feldman2003visual, palmeri2004visual, scholl2001visual, cavanagh2023architecture}. Our model allows us to systematically vary the input sources (e.g., semantic, motion-based, or appearance-based object cues) used for the formation of the scene segmentation, which defines the perceptual units on which the object-based attentional selection process operates. 
Under the assumption that our implementation of saccadic decision-making mechanisms is similar to the human visual system, we expect that the object cues that lead to more human-like scanpaths are also the cues primarily used for saccadic decision-making in humans.

Our results suggest that attentional guidance primarily relies on semantic object cues in dynamic scenes. Only models that used the semantic cues both for the global and prompted scene segmentation showed the temporal IOR effect and could reproduce the balance between foveation categories seen in humans (cf.~\Cref{fig:object_ablation}). 
This result is consistent with evidence for global semantic understanding of natural scenes \citep{neri2017object,cavanagh2023architecture}. 
As expected, the model scanpaths also became less human-like if we replaced the prompted semantic segmentation at the gaze position with an appearance-based, low-level object cue prompted at the fixation position (\emph{all-g \& ll-p} in \Cref{fig:object_ablation}, overestimating the amount of inspection and not showing the temporal IOR effect). This model corresponds to the assumption that a foveated object would get more finely segmented (e.g., a t-shirt, which was previously part of a person, becomes its own object when foveated). However, we do not see support for this assumption since the simulated scanpaths based on it were less plausible compared to the base model. Removing the global low-level object cues (\emph{hl-g \& hl-p} in \Cref{fig:object_ablation}) did not impact the simulated scanpath statistics in any major way. 
There is ample evidence for the brain using appearance- and motion-based object cues to segment complex dynamic scenes \citep{schyns1994blobs, reppas1997representation, vonderheydt2015figure}. Based on our results, however, we would argue that low-level object cues do not play an important role in the formation of the perceptual units on which object-based attention is operating.

These results could be tested experimentally by probing the visual sensitivity within or outside the currently foveated object as predicted by the model. A promising method to study this would be to test the response to gaze-contingent narrow-band contrast increments during free viewing \citep{dorr2013peri}. Under the assumption of a delayed response to probes outside an attended object \citep{egly1994shifting, scholl2001objects} and in combination with the predictions from our model, this would allow us to disentangle the object cues used in the visual system to construct perceptual units for object-based attention.

\subsection{Employing an information processing pattern from robotics}\label{sec:dis_aicon}
Our model is based on the robotics-inspired information processing pattern Active InterCONnect (AICON), which structures information processing at a mechanistic level to generate adaptive behavior. Our results and recent studies show that AICON is not limited to robotics but is applicable to domains like human perception of visual illusions~\citep{battaje2024aicon} or even collective behavior~\citep{mengers2024leveraging}, where systems must integrate uncertain, interdependent inputs to make perceptual decisions. Here, we present our evidence for how AICON's algorithmic patterns address the specific challenges of human vision, which show strong parallels to those in robotics. Based on this evidence, we then provide a “recipe” for building AICON-ic models of other perceptual processes.

Building a model with AICON means constructing a system of recursive components that interact through actively modulated bidirectional connections (\emph{active interconnections}). As discussed for our model in \Cref{sec:dis_study}, there is ample evidence for recursive updating in human perception. These recursions within perceptual processes---often implemented in a Bayesian way---are critical for resolving ambiguous inputs, whether from sensory neurons or a robot’s camera. For example, recursive processing turns depth perception, a nearly impossible task when attempted with a single image, into a trivial one by incorporating motion parallax. Active interconnections between components further refine perception, since they can share relevant information extracted through other means with each other. In integrating cues in this way, perception becomes more robust, as seen in robotics~\citep{martin2022coupled,mengers2023combining}. But this goes beyond simple one-directional cue integration: Because each recursive component remains uncertain, it should use all available information from its related components to reduce its uncertainty. Active interconnections are \emph{bidirectional} and the conveyed information between components needs to adapt to changing uncertainties to ensure an information flow from more to less certain components at all times (\emph{active modulation}). In our model, the active interconnection between the object segmentation and saccadic decision-making module leads to human-like visual exploration behavior. Likewise, \citet{battaje2024aicon} have shown that active interconnections between color and shape perception and between luminance and motion perception enable models to replicate the human perception of visual illusions while accounting for individual variability.

We believe AICON will be transferable to other vision processes. For those interested in building AICON-inspired models, we offer both our code (on GitHub: \url{https://github.com/rederoth/AICONic_ScanDy}) and suggest a general three-step recipe: (1) Identify key perceptual processes or representations likely to contribute to the high-level process of interest. Build a recursive model for each, ideally one that estimates uncertainty over its representation, and verify each component’s behavior in isolation using controlled inputs. (2) Define and implement active interconnections between components based on possible interdependencies, modulating these connections based on component states and uncertainties. Add connections incrementally, observing and tuning system behavior to align with expected outcomes. (3) Once fully connected, observe how behaviors emerge from component interactions. Experiment with ablating connections or adjusting parameters to refine alignment with experimental data or to generate new predictions, such as individual variability in perceptual processes.

By applying this recipe, AICON offers a versatile framework that fosters knowledge exchange across disciplines studying behavior, like robotics and vision science. Although behaviors are very different on a lower level (how computation is exactly performed) and a higher level (the exact ecological niche and its constraints), the common mechanistic challenges---integrating uncertain information and adapting across contexts---often result in convergent solutions. Therefore, we believe that studying mechanistic information processing patterns like AICON across disciplines offers a promising path toward a more unified and deeper understanding of the fundamental drivers of behavior.

\section{Conclusion}
\label{sec:conclusion}

We developed and evaluated a model for object-based attention and gaze behavior in real-world dynamic scenes. By integrating saccadic decision-making mechanisms with an object segmentation framework, our model successfully simulates human-like scanpaths. This integration, an implementation of the Active InterCONnect (AICON) information processing pattern from robotics, enables the model to progressively refine its object segmentation through active exploration, while uncertainty over that segmentation guides the scanpath.

The modular design of our model allows for systematic hypothesis testing and ablation studies, providing a valuable tool for exploring the mechanisms of visual attention. We found that the uncertainty in object segmentation plays a crucial role in guiding human-like visual exploration. Instead of relying on an explicit ``inhibition of return'' (IOR) mechanism, we propose the active reduction of uncertainty through saccadic decisions as the driving mechanism of scene exploration.
Furthermore, our results suggest that attentional guidance primarily relies on semantic object cues, highlighting the importance of high-level scene understanding in active vision.
By capturing the interplay of segmentation and saccadic decision-making, our model highlights the power of mechanistic information processing patterns like AICON, encouraging future research to explore information processing patterns that transcend disciplinary boundaries.

\section*{Acknowledgements}
We thank Richard Schweitzer, Madeleine Gross, and Olga Shurygina for their permission to use the UVO eye-tracking dataset and Julie Ouerfelli-Ethier for her helpful comments on the manuscript. We gratefully acknowledge funding by the Deutsche Forschungsgemeinschaft (DFG, German Research Foundation) under Germany's Excellence Strategy---EXC 2002/1 ``Science of Intelligence''---project number 390523135.

\bibliographystyle{plainnat}
\bibliography{refs}

\newpage
\appendix

\section{Uncertainty over object segmentation} \label{app:uncertainty}

\begin{figure}
    \centering
    \includegraphics[width=1\linewidth]{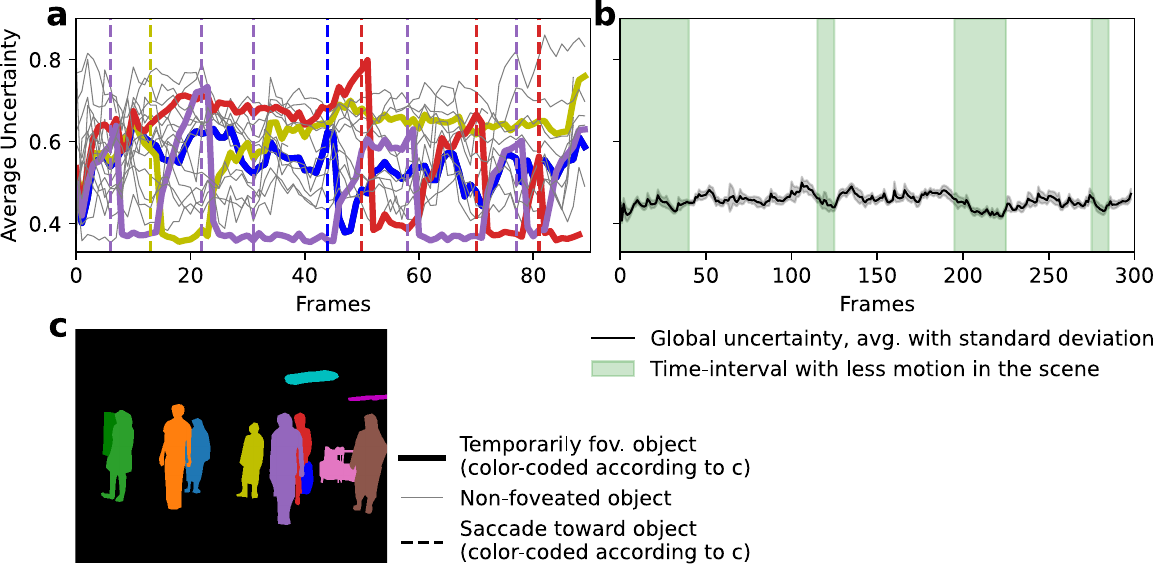}
    \caption{Uncertainty in our model represents where the boundary of objects is currently ambiguous. We visualize the uncertainty $U'$ of our model for the same scene as in \Cref{fig:qualitative}. We show the uncertainty for individual ground truth objects in (a) with the ground truth objects of that scene shown for the initial frame in (c). The uncertainty of non-foveated objects (thin gray lines) varies for different objects and over time, but after a saccade towards an object (dashed colored line indicates a saccade towards the same color object) uncertainty for that object (thick colored line) rapidly reduces and remains low until the gaze moves away. (b) Global uncertainty (averaged over 10 scanpaths with standard deviation) remains in the same regime but reduces if there is less motion in the scene (green time intervals), since ambiguity can be resolved while less new ambiguity arises from motion.}
    \label{fig:uncertainty_intuition}
\end{figure}

We estimate the uncertainty over object segmentation using a particle filter, as described in \Cref{sec:mandm_segmentation}. This uncertainty reflects the ambiguity in the existence and location of boundaries between objects and not the identity or category of individual objects. Given its key role in our model, we provide an intuitive explanation of this uncertainty and how it is updated on a frame-by-frame basis, illustrated with an example sequence in \Cref{fig:qualitative}. We visualize the model’s uncertainty $U'$ both as an average across the entire scene (\Cref{fig:uncertainty_intuition}b) and as an average over the ground truth objects in the video (\Cref{fig:uncertainty_intuition}a, with the ground truth objects shown in \Cref{fig:uncertainty_intuition}c).
For individual ground truth objects, we show how gaze position affects object uncertainty for a single simulated scanpath over the first 90 frames with available ground truth. For the global uncertainty across the scene, we average over 10 stochastic scanpath realizations to observe the general relationship between uncertainty and scene content, independent of the specific scanpath.

After the initial pre-attentive global segmentation, the uncertainty for different objects varies based on visual ambiguity in appearance and semantic cues (cf. \Cref{fig:uncertainty_intuition}a). Uncertainty for non-foveated objects depends on factors like motion or occlusion. However, if an object is foveated (e.g., the red object at frame 50), its uncertainty rapidly decreases to nearly the minimum ($u_\mathrm{min}=\nicefrac{1}{3}$). When gaze shifts away from this object (e.g., in frame 58), its uncertainty rises again. This dependency on various factors, combined with diverse uncertainties across objects, means that average uncertainty over the scene remains relatively steady throughout the sequence (cf.~\Cref{fig:uncertainty_intuition}b). This stability arises because we are not estimating uncertainty over object identity, which would remain low once identified, but over object boundaries, which can quickly become ambiguous again as objects move, their visible parts change, or occlusions occur. If there is no camera movement and the objects in the scene remain mostly static, the overall uncertainty decreases (cf.~the green intervals in \Cref{fig:uncertainty_intuition}b, where dancers either slowly rotate or have a stop-step and thus move less).

\begin{figure}
    \centering
    \includegraphics[width=1\linewidth]{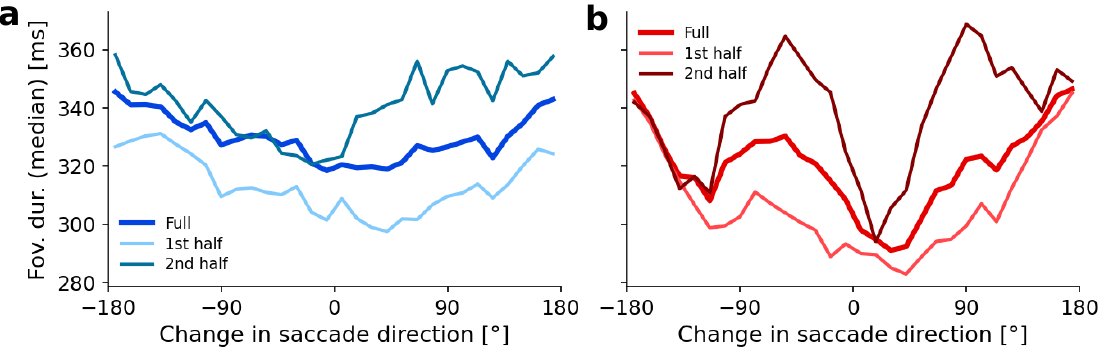}
    \caption{Temporal IOR effect during the first and second half of presentation time. (a) Evaluation of the base model with the result of the full duration, as shown in \Cref{fig:base_stats}d, compared to an evaluation of foveations only during the first (frames 0-149) or second half (frames 150-299). (b) Same as (a) for human data.}
    \label{fig:uncertainty_ior_timebins}
\end{figure}

As a result, the global uncertainty remains on a similar level over time as long as there are ongoing scene changes in the video. Hence, we did not expect a strong effect on the temporal development of uncertainty-related effects. To test this, we compared the temporal IOR effect for the first and second half of the videos (frames 0-149 and frames 150-299) for both the base model and the human data, as shown in \Cref{fig:uncertainty_ior_timebins}. The general trend of a temporal IOR effect remains visible for both halves, only with higher variability due to the smaller number of samples. However, the average foveation durations are shorter for the first half of frames than for the second for both our model and the human data. In the model, this must be the result of changes in uncertainty, since the salient feature map $F'$ is always normalized and the visual sensitivity $S$ does not change in magnitude. However, since the global uncertainty also does not change drastically, this is most likely a result of the distribution of uncertainty compared to the distribution of saliency. The first saccades favor objects with both high saliency and uncertainty, leading to a reduction of uncertainty for these objects. Due to this reduced uncertainty, evidence for these objects that initially drive a fast sequence of saccades is then accumulated more slowly, leading to overall longer foveation durations.

In summary, uncertainty in our model represents where the boundary of objects is currently ambiguous. This can fluctuate depending on the scene content, but by foveating on an object, the model ensures a lower uncertainty for that object. However, uncertainty for objects can rise after gaze shifts away, leading to a relatively stable amount of global uncertainty over the scene that distributes according to scanpath history and scene content over time.

\section{Semantic and foveated segmentation from prompt-based models}\label{app:semanticsegmentation}

We use a state-of-the-art data-driven model~\citep{kirillov2023sam} to generate both the pre-attentive global segmentations and the segmentations of the currently foveated object, as described in \Cref{sec:mandm_segmentation_cues}. This family of models is fundamentally based on the formulation of segmentation as a prompt-based task. Here, we give a high-level explanation of this task formulation and how such models are trained with vast amounts of data to provide an intuition on how the semantic object cues were obtained. For further details on the general concept, refer to~\citet{kirillov2023sam}, and for details on the specific derived models used here, see~\citet{ke2023samhq} and~\citet{zhao2023fastsam}.

Traditionally in computer vision, segmentation has been formulated as the task of generating a dense map for a given image that separates it into regions, often based on known object classes. \citet{kirillov2023sam} introduced an alternative formulation in which, given a prompt relating to one object in the image (a point, a bounding box, text, or a dense mask), the mask covering that object needs to be identified. For this formulation of the segmentation task, a learnable model consists of encoders for the image as well as all possible prompts and a decoder that, given the encoded image and prompt, generates a mask. These en- and decoders are usually transformer-based and can then be trained in unison based on a vast amount of data of labeled segmentations collected from various sources~\citep{kirillov2023sam}. The new task formulation, moreover, allows leveraging labeled segmentations in multiple ways, since one image can be used for all types of prompts, as well as variations of the same prompt, e.g., by shifting the prompt point within the labeled mask of that object. Thus, the amount of training samples increases and a model that can solve a variety of segmentation tasks can be learned.

We use such learned models with prompt-points to generate both pre-attentive global semantic segmentations and the segmentations of the currently foveated object. Let us give an example based on \Cref{fig:modeloverview} starting with the more intuitive segmentation of the foveated object: the current gaze acts as prompt-point that currently lies on the face of the person in the foreground, and thus an appropriate mask (the segmentation of the foveated object) contains the entire person. Note that alternative appropriate masks might only be the person's face or maybe even only the specific part of the face the prompt is on, e.g., the left eye. Hence, models for this task do not output just one mask, but a weighted set of masks from which the user can select (we simply use the highest weighted mask throughout this work). 
This segmentation procedure is in itself not foveated, as it leverages the entire image in a spatially invariant manner during encoding and only leverages the prompt-point during the final mask construction. To generate a global semantic segmentation of the scene with such a model, instead of passing one point, we pass a grid of points to obtain a set of masks that can be combined into one segmentation (cf. the semantic segmentation in \Cref{fig:object_seg_model}).

\section{Further details on the particle filter implementation}\label{app:particle_filter}
We track a belief over scene segmentation by combining different measurements over time within a particle filter, as we describe in \Cref{sec:mandm_segmentation}. Here, we want to give further details on its implementation, especially regarding the computation of each particle's weight and the matching process for segmentation IDs when marginalizing the particle set into a single object segmentation.
\subsection{Particle weighting} \label{app:particle_weighting}
When computing the weight of each particle (\Cref{eq:particle_filter_weight}), we weigh it according to each segmentation cue. To do so, we first compute the unnormalized weights $\widetilde{w}^{[\mathrm{i}]}_\mathrm{t}(z_\mathrm{t})$ according to each cue $z_\mathrm{t}$, using a distance function between two segmentations (\Cref{eq:weight_dist}). We can determine this distance between two segmentations as the sum of the distances of each boundary
pixel in one segmentation to the closest boundary
pixel in the other, which is easily computable using the distance transform $\mathrm{disttransform}(s)$ of the boundary image $s$ of a segmentation. Since this distance is non-symmetric, we, however, need to use it in both directions (\Cref{eq:dist_seg} where $W$ and $H$ are the width and height of the image frame).

\begin{gather}
    \widetilde{w}^{[\mathrm{i}]}_\mathrm{t}(z_\mathrm{t}) = \frac{1}{d(s^{[\mathrm{i}]}_\mathrm{t}, z_\mathrm{t})}\label{eq:weight_dist}\\
    \begin{split}d(s_1, s_2) = \sum_{x=1}^W \sum_{y=1}^H \Big(&(s_1)_{xy} \cdot (\mathrm{disttransform}(s_2))_{xy}\\
        &+ (s_2)_{xy} \cdot (\mathrm{disttransform}(s_1))_{xy}
    \Big)\end{split}\label{eq:dist_seg}
\end{gather}

To determine the overall weight of a particle according to the set of all cues $\mathcal{Z}_\mathrm{t}$ at time $t$, we combine the unnormalized weights $\widetilde{w}^{[\mathrm{i}]}_\mathrm{t}(z_\mathrm{t})$ for each cue $z_\mathrm{t}$ as in a product, as shown in \Cref{eq:final_particle_weight} where $\eta$ is a normalizing factor between particles. However, as the cues have different amounts of information and thus confidence, we combine with an additional exponential importance factor $\alpha_\mathrm{z}$. These importance factors were set during some initial explorations on the dataset to produce satisfactory segmentations as shown in \Cref{tab:params}.
\begin{equation}
    w^{[\mathrm{i}]}_\mathrm{t}(\mathcal{Z}_\mathrm{t}) = \frac{1}{\eta} \prod_{z_\mathrm{t} \in \mathcal{Z}_\mathrm{t}} \big( \widetilde{w}^{[\mathrm{i}]}_\mathrm{t}(z_\mathrm{t}) \big)^{\alpha_\mathrm{z}}\label{eq:final_particle_weight}
\end{equation}

\subsection{Matching segmentation IDs for consistency over time}\label{app:id_matching}
 
We obtain a single segmentation from the particle set during each iteration to inform saccadic decision-making, as we have described in \Cref{sec:mandm_segmentation}. To keep the IDs of this segmentation consistent, we use a variation of the Hungarian algorithm~\citep{hopcroft1973} to match object IDs between object segmentations. To do so, we must determine the matching weights $w_\mathrm{m}(m_1, m_2)$ between the mask $m_1$ of an object in one segmentation and the mask $m_2$ in another. We use the well-established Intersection over Union $\mathrm{IOU}(m_1, m_2)$ metric to measure their overlap:
\begin{equation}
    \mathrm{IOU}(m_1, m_2) = \frac{m_1 \cap m_2}{m_1 \cup m_2}~.
\end{equation}

However, if we only consider these overlaps between the current and last segmentation, some object IDs will get lost due to perceptual uncertainty. Hence, we consider the last 10 segmentations but discount their importance with the factor $\beta$. We compute resulting matching weights $w_\mathrm{m}(m_1, m_2)$ that the mask $m_1$ in the current segmentation should have the same ID as the mask $m_{2}$ in each of the last $T=10$ segmentations following
\begin{equation}
    w_\mathrm{m}(m_1, m_2) = \sum_{t=0}^{T} \mathrm{IOU}(m_1, m_{2, (T - t)}) \cdot \beta^{(T - t)}~,
\end{equation}
and then match the IDs using maximum weight full matching in bipartite graphs~\citep{jonker1988shortest}, allowing for new IDs if no existing ID can be matched.

\section{Parameter exploration}
\label{app:parameter_exploration}

We found appropriate parameter values through extensive grid searches in a four-dimensional parameter space, as described in \Cref{sec:mandm_fitting}.
To make the computational cost of the grid search feasible, we fixed all parameters except for the decision threshold $\theta$ for the drift-diffusion model (DDM), the DDM noise level $s$, and the scaling parameters for the importance of the uncertainty $u_\mathrm{min}$ and salient scene features $f_\mathrm{min}$. All free and fixed model parameters are described in \Cref{tab:params}.

\begin{figure}
    \centering
    \includegraphics[width=0.8\linewidth]{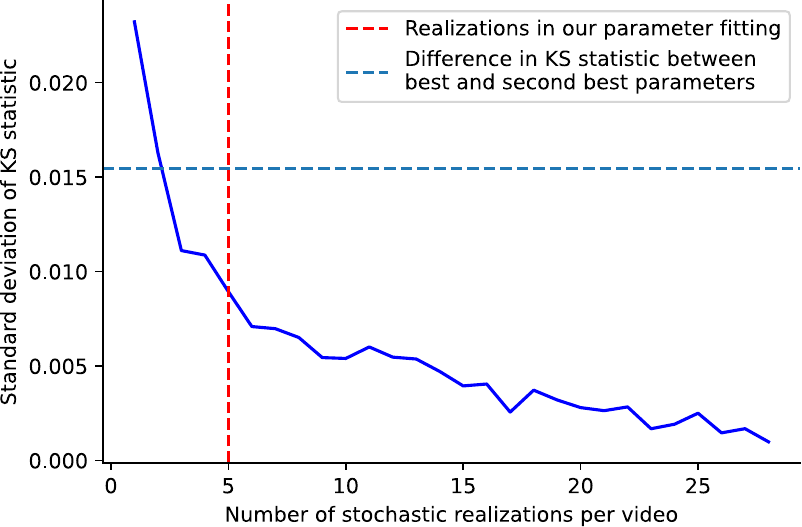}
    \caption{To ensure a reliable estimation of the model's free parameters on the training set, we compared the variability of the KS statistic for the base model across different numbers of stochastic scanpath realizations to the difference between the two best-fitting parameter sets. Initially, adding more realizations significantly reduced the standard deviation, bringing it below the difference in KS statistic between the best-fitting parameter sets. Beyond this point, additional realizations only gradually reduced variability. Therefore, we selected 5 stochastic scanpath realizations per video to fit our model parameters.}
    \label{fig:ks_stats}
\end{figure}

We used 10 videos from our dataset as a training set (33 for testing) and generated 5 scanpaths stochastically for each video and parameter configuration. To confirm that 5 scanpath realizations were sufficient to estimate the model’s free parameters reliably, we assessed the variability of the KS statistic over different numbers of realizations. For this, we simulated for each video in the training set 30 scanpath realizations using the base model parameters. For each number $N\in[2,29]$, we then randomly drew 25 sets of size N out of the 30 realizations and calculated the KS statistic for each set. The standard deviation of the KS statistic across the 25 sets for each number of stochastic realizations $N$ is shown in \Cref{fig:ks_stats}. We used the difference in KS statistic between the two best-fitting parameter sets as a reference value and found that with 5 realizations, the KS statistic's standard deviation was already below that difference. Increasing the number of realizations beyond 5 only slightly reduced variability. Therefore, we used 5 stochastic scanpath realizations to reduce the computational cost of the grid search. The model’s strong generalization from training to test set throughout our results validated this approach.

\begin{table}[p]
    \centering
    \begin{tabularx}{\textwidth}{|l|X|}
        \hline
         $\mbox{\boldmath$\theta$}=4.0$ &  Decision threshold of the drift-diffusion model.  \\\hline
         $\mathbf{s}=0.4$& Decision noise of the drift-diffusion model.\\\hline
        $\mathbf{u}_\mathrm{\textbf{min}}=\nicefrac{1}{3}$  & Parameter to rescale the uncertainty map $U$ to $U'\in [u_\mathrm{min},1]$. Increasing $u_\mathrm{min}$, reduces the importance of uncertainty in our model.\\\hline
        $\mathbf{f}_\mathrm{\textbf{min}}=0$  & Parameter to rescale the salience map $F$ to $F'\in [f_\mathrm{min},1]$. Increasing $f_\mathrm{min}$, reduces the importance of salience in our model.\\\hline
        $\sigma_\mathrm{S} = 7$ dva & Standard deviation of the isotropic Gaussian $G_\mathrm{S} = \frac{1}{2\pi\sigma_\mathrm{S}^2}\exp\left(-\frac{(x-x_0)^2+(y-y_0)^2}{2\sigma_\mathrm{S}^2}\right)$ that we use to model visual sensitivity. \\\hline
         $N=50$ & Number of segmentation particles. Higher number improves estimation of both segmentation and uncertainty, but heavily increases computational load. We find that $N=50$ is already sufficient, given the direct insertion of regions to prevent divergence.\\\hline
         $\alpha_\mathrm{appearance}=0.4$ & Importance factor of the appearance segmentation to determine the weight of each particle. \\\hline
          $\alpha_\mathrm{motion}=0.05$ & Importance factor of the motion segmentation to determine the weight of each particle. It is set much lower than others since the motion segmentation only provides information about some parts of the scene, while also being noisy.\\\hline
        $\alpha_\mathrm{semantic}=1.0$ & Importance factor of the semantic segmentation to determine the weight of each particle. \\\hline
        $\alpha_\mathrm{foveated}=0.6$ & Importance factor of the foveated segmentation to determine the weight of each particle. It is lower than for the semantic segmentation, since it only provides information on part of the scene, while having in principle the highest confidence.\\\hline
       $r_\mathrm{scale, foveated}=1.0$ & Scale factor of the input resolution for the foveated object segmentation. \\\hline
      $r_\mathrm{scale, other}=0.35$ & Scale factor of the input resolution for all other segmentation cues. They are downsampled to model lower-confidence information. \\\hline
    \end{tabularx}
    \caption{Parameters of our model. We show their settings for our base model, with the parameters that are fitted for different versions of the model in \textbf{bold}.}
    \label{tab:params}
\end{table}

\begin{figure}[p]
    \centering
    \includegraphics[width=0.95\linewidth]{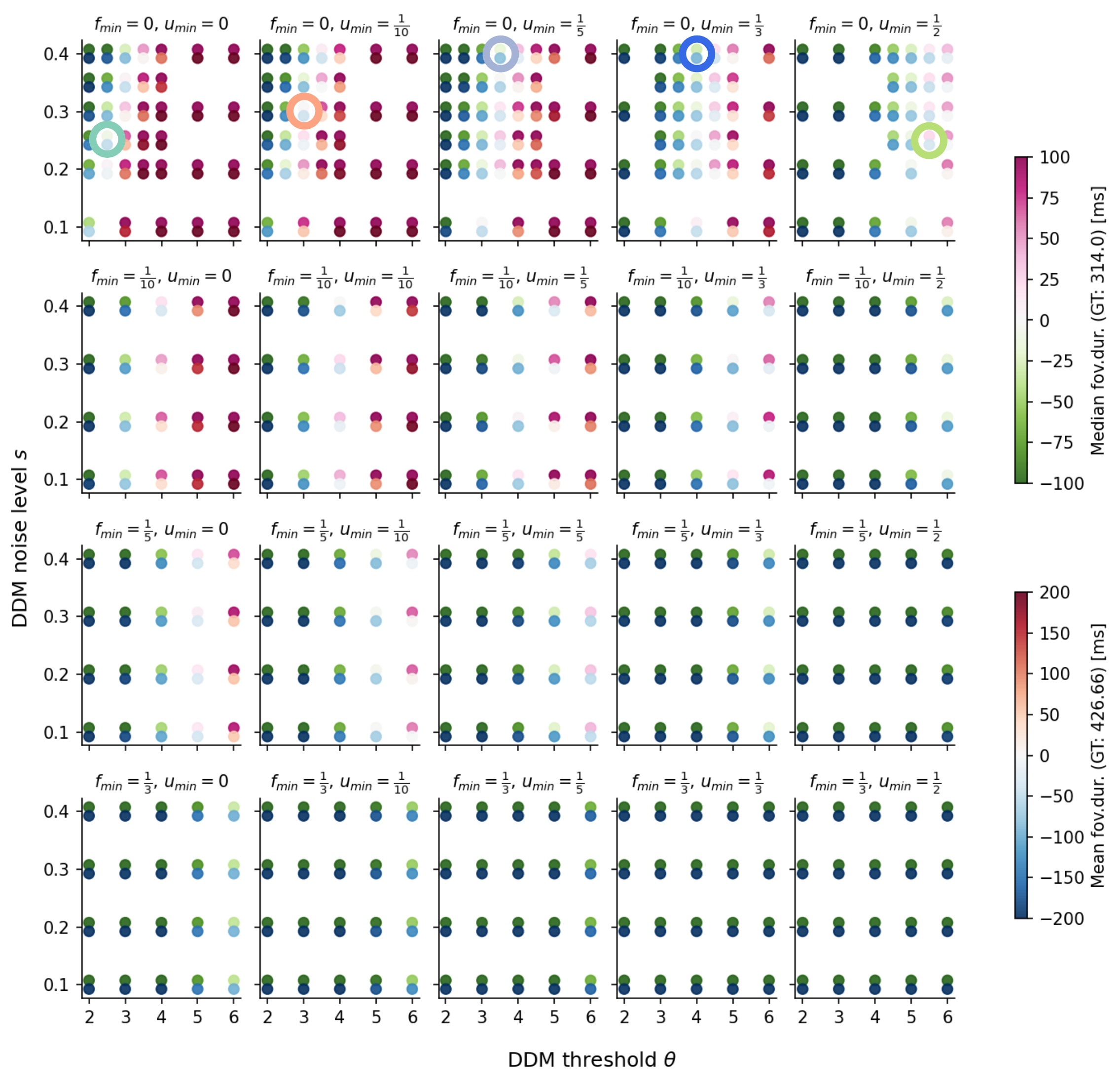}
    \caption{The distribution of foveation durations is one criterion to determine model parameters within the 4-dimensional grid of free parameters. Each dot-tuple characterizes the deviation of the median (higher) and mean (lower dot) foveation duration of simulated scanpaths compared to the human ground truth (GT) in the training set (10 videos; 5 random seeds each). Brighter dots indicate more suitable parameters. Circles mark the chosen parameter sets for each value of $u_\mathrm{min}$, which we subsequently analyzed in detail as shown in \Cref{fig:uncert_stats}.}
    \label{fig:fov_dur_grid}
\end{figure}
\begin{figure}[p]
    \centering
    \includegraphics[width=1\linewidth]{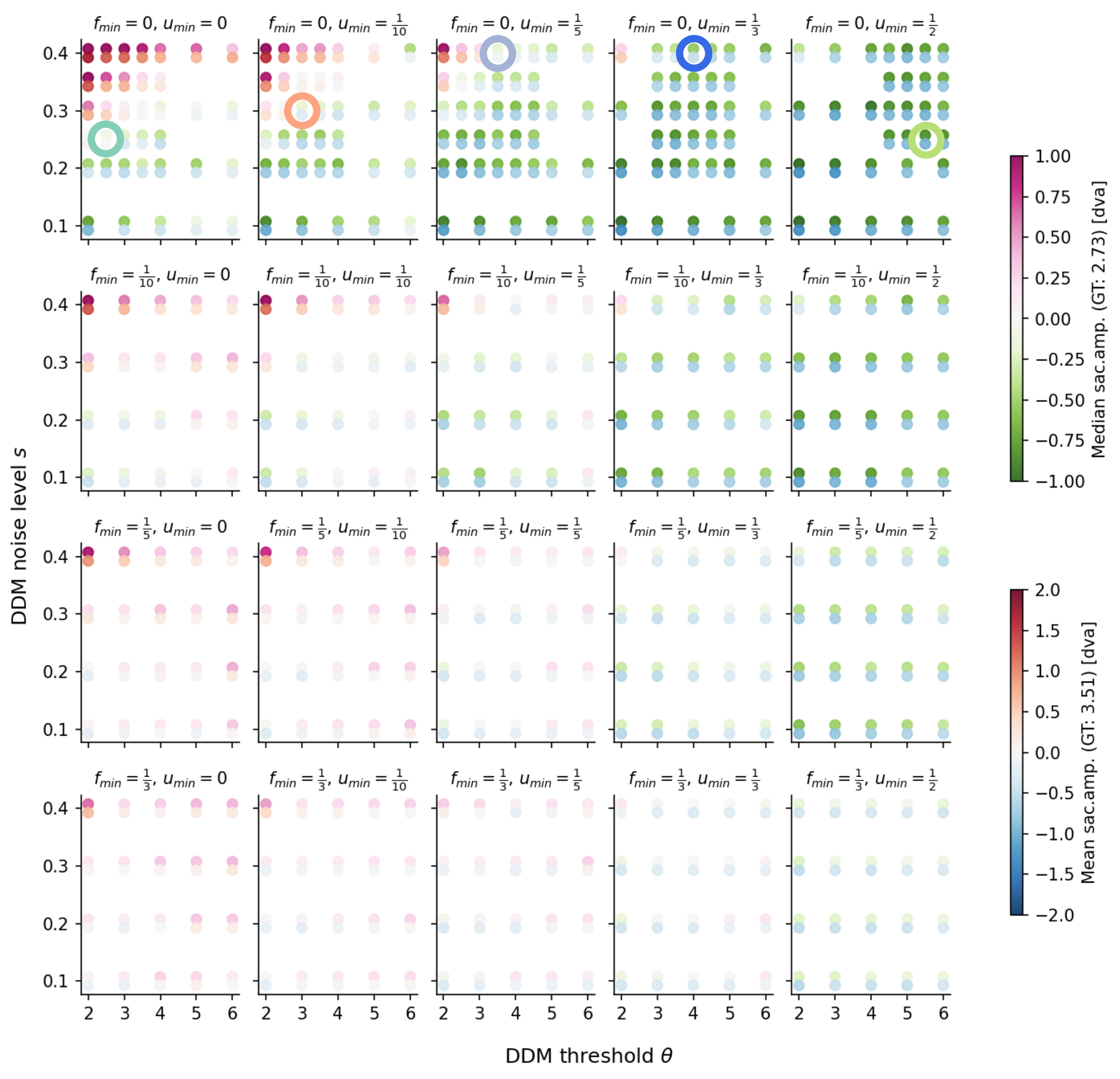}
    \caption{The saccade amplitude distribution of the simulated scanpaths is the second criterion, plotted analogously to \Cref{fig:fov_dur_grid}.}
    \label{fig:sac_amp_grid}
\end{figure}

We first ran a coarse parameter grid exploration for the parameters $\theta\in[2,3,4,5,6]$, $s\in[0.1,0.2,0.3,0.4]$, $u_\mathrm{min}\in[0,\frac{1}{10},\frac{1}{5},\frac{1}{3},\frac{1}{2}]$, and $f_\mathrm{min}\in[0,\frac{1}{10},\frac{1}{5},\frac{1}{3}]$. Around the best-performing parameters, we performed a finer grid search in $\theta$ and $s$, as shown in Figs.~\ref{fig:fov_dur_grid} and~\ref{fig:sac_amp_grid}. 
We did not consider parameter sets with $s>0.4$ since previous model explorations have shown that the simulated scanpaths for higher noise levels are more likely to explore the background or objects that are not often foveated by human observers. As the main indication for noise-driven scanpaths, we took a lower correlation of the object dwell time between simulated and human scanpaths, as it is shown in \Cref{fig:extension_showcase}b, which in fact decreases for models with $s>0.4$.

To ensure a fair comparison between models in our ablation studies, we run additional parameter explorations for the models where the foveation duration and saccade amplitude change considerably compared to the base model with the parameters in \Cref{tab:params}. 
For the model without uncertainty contribution (\emph{no uncert.} in \Cref{fig:uncert_stats}) we set $U'=u_\mathrm{min}$, resulting in a model with $\theta=3.0, s=0.3, f_\mathrm{min}=0, u_\mathrm{min}=\frac{1}{3}$ having the lowest mean of KS statistics $D_\mathrm{FD}$ and $D_\mathrm{SA}$ across the 4-dimensional grid of free parameters.
When investigating the influence of different object cues, we explored a fine parameter grid $f_\mathrm{min}=0, u_\mathrm{min}=\frac{1}{3}$ for better comparability with the other models. This resulted in parameter values of $\theta= 4.0, s=0.4$ for the model using ground truth objects (\emph{gt-obj} in \Cref{fig:object_ablation}), $\theta=5.5, s=0.4$ for the model with all global object cues but without a prompted object (\emph{all-g \& no-p}), and $\theta=5.5, s=0.4$ for the model with global appearance and motion-based segmentation only (\emph{ll-g \& no-p}). 

\section{Videos of human and model scanpaths}\label{app:videos}
The visualizations of our model parts shown in \Cref{fig:modeloverview,fig:object_seg_model,fig:scanpath_model,fig:qualitative} can be seen as \href{https://doi.org/10.14279/depositonce-22812}{downloadable} videos for $10$ different simulated scanpaths on that input sequence. We show $10$ simulated scanpaths for $10$ additional videos from the test set to illustrate the variability of our dataset.
For comparison, we also show the scanpaths of $10$ human participants on the respective input sequences. 
All videos are shown with a playback speed of 0.5 (i.e., 15 instead of 30 fps) to make it easier to compare the scanpaths.

\section{Extended models: Details and statistics}
\label{app:extensions}

\begin{figure}
    \centering
    \includegraphics[width=1\linewidth]{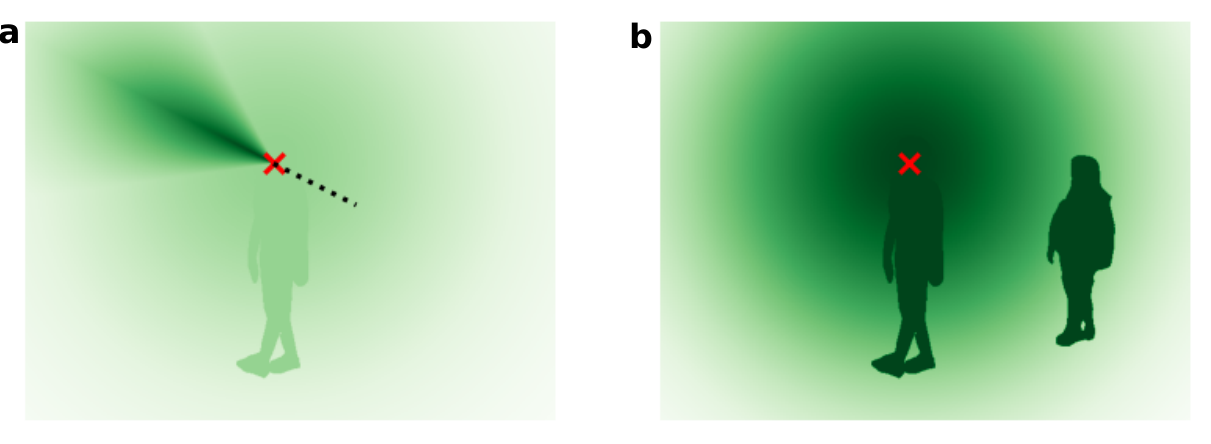}
    \caption{Illustration of the modified sensitivity maps $S'$ for the two extended models. (a) Saccadic momentum: We set the maximal value in the direction of the previous saccade (indicated with the dotted line) to $2.5$, which decreases linearly to $0.85$ within an angle of $35^\circ$, and multiply the resulting map with $S$. (b) Pre-saccadic attention: If the evidence of an object crossed 30\% of the decision threshold $\theta$, we obtain a prompted object mask at its location and set the sensitivity of this object to 1.}
    \label{fig:extension_sensitivity}
\end{figure}

\begin{figure}[t]
    \centering
    \includegraphics[width=1\linewidth]{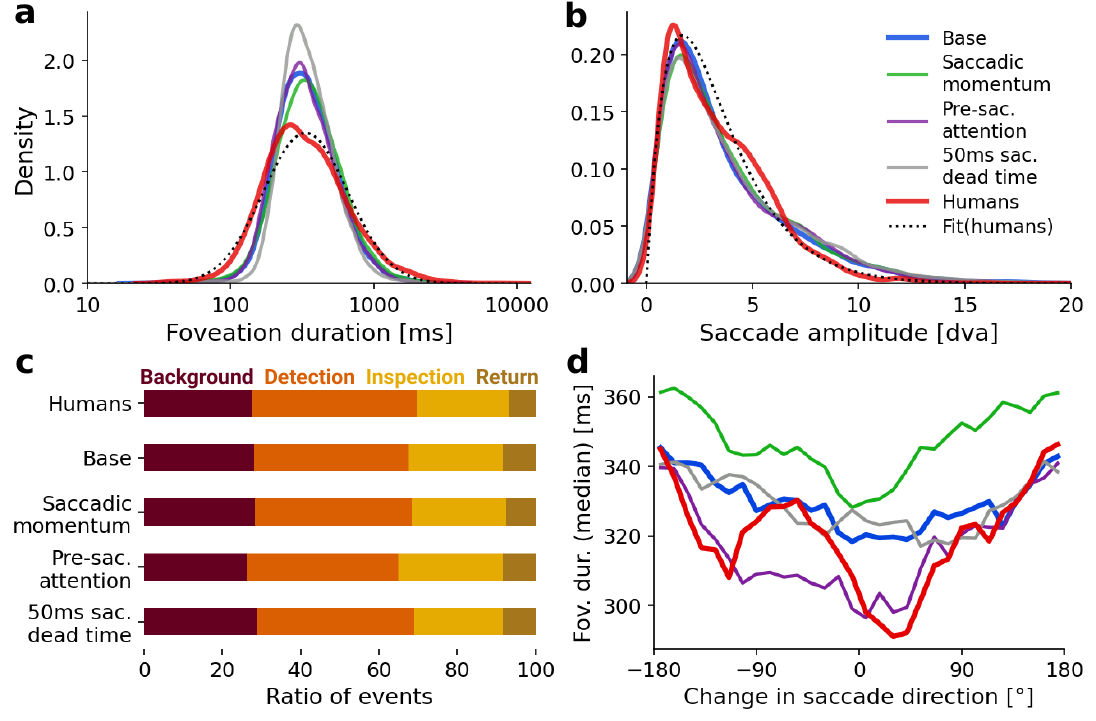}
    \caption{Aggregated scanpath statistic of the extended models, analogous to \Cref{fig:uncert_stats}.}
    \label{fig:extension_stats}
\end{figure}


\end{document}